\documentclass[preprint,authoryear,3p]{elsarticle}

\usepackage{amssymb,amsthm,amsmath}

\usepackage{graphicx}
\usepackage{subfig}
\usepackage{float}
\usepackage{multirow}
\usepackage{longtable,lscape}
\usepackage{cases}
\usepackage{url}
\usepackage{caption}
\usepackage{color}
\usepackage{amsthm}
\usepackage{algpseudocode,algorithm,algorithmicx}
\usepackage[pagewise]{lineno}
\usepackage{comment}
\usepackage{graphicx}
\usepackage{rotating}
\usepackage{bbm}

\usepackage{enumitem}
\DeclareMathOperator*{\argmax}{arg\,max}  

\usepackage{fancyhdr}
\usepackage{fullpage}

\theoremstyle {plain}

\newcounter{x}

\theoremstyle{definition}

\newtheorem{exmp}{Example}[section]

\theoremstyle{remark}
\newtheorem*{rem}{Remark}
\newtheorem*{note}{Note}

\usepackage{titlesec}
\titleformat{\paragraph}
{\normalfont\normalsize}{\theparagraph}{1em}{}
\titlespacing*{\paragraph}
{0pt}{3.25ex plus 1ex minus .2ex}{1.5ex plus .2ex}

\usepackage{color}

\usepackage{lipsum}

\def\ps@pprintTitle{%
	\let\@oddhead\@empty
	\let\@evenhead\@empty
	\def\@oddfoot{\reset@font\hfil\thepage\hfil}
	\let\@evenfoot\@oddfoot
}
\makeatother

\makeatletter
\def\ps@pprintTitle{%
	\let\@oddhead\@empty
	\let\@evenhead\@empty
	\let\@oddfoot\@empty
	\let\@evenfoot\@oddfoot
}
\makeatother

\usepackage{multicol}

\journal{Transportation Research Part C}

\begin{document}


\begin{frontmatter} 
		
		
		
\title{Reward Design for Driver Repositioning Using Multi-Agent Reinforcement Learning}

\noindent \textcolor{blue}{Published in: Transportation Research Part C 119 (2020) 102738.}

\noindent \textcolor{blue}{Please cite this Paper as: Shou, Z., Di, X., 2020. Reward design for driver repositioning using multi-agent reinforcement learning. Transportation Research Part C: Emerging Technologies 119, 102738. DOI: 10.1016/j.trc.2020.102738}
		
		
		
\author[cu]{Zhenyu Shou}

\author[cu,dsi]{Xuan Di\corref{cor}}
\ead{sharon.di@columbia.edu}

\cortext[cor]{Corresponding author. Tel.: +1 212 853 0435;}

\address[cu]{Department of Civil Engineering and Engineering Mechanics, Columbia University}
\address[dsi]{Data Science Institute, Columbia University}

\begin{abstract}

A large portion of passenger requests is reportedly unserviced, partially due to vacant for-hire drivers' cruising behavior during the passenger seeking process. This paper aims to model the multi-driver repositioning task through a mean field multi-agent reinforcement learning (MARL) approach that captures competition among multiple agents.
Because the direct application of MARL to the multi-driver system under a given reward mechanism will likely yield a suboptimal equilibrium due to the selfishness of drivers, this study proposes a reward design scheme with which a more desired equilibrium can be reached. To effectively solve the bilevel optimization problem with upper level as the reward design and the lower level as a multi-agent system, a Bayesian optimization (BO) algorithm is adopted to speed up the learning process.  
We then apply the bilevel optimization model to two case studies, namely, e-hailing driver repositioning under service charge and multiclass taxi driver repositioning under NYC congestion pricing. 
In the first case study, the model is validated by the agreement between the derived optimal control from BO and that from an analytical solution. With a simple piecewise linear service charge, the objective of the e-hailing platform can be increased by $8.4\%$. 
In the second case study, an optimal toll charge of $\$5.1$ is solved using BO, which improves the objective of city planners by $7.9\%$, compared to that without any toll charge. 
Under this optimal toll charge, the number of taxis in the NYC central business district is decreased, indicating a better traffic condition, 
without substantially increasing the crowdedness of the subway system. 


\end{abstract}

\end{frontmatter}


\section{Introduction}

The emergence of transportation network companies (TNCs) or e-hailing platforms (such as Didi and Uber) has revolutionizsed the traditional taxi market and provided commuters a flexible-route door-to-door mobility service. Nonetheless, it is reported that a large portion of the passenger requests remain unserviced \citep{lin_efficient_2018}, partly due to the imbalance between demand (i.e., passenger requests) and supply (i.e., available drivers) 
that results in long cruising trips for taxi drivers to  
find the next passenger 
and long waiting time for passengers to be picked up \citep{powell_towards_2011,di2019rue_e}. 
Such cruising behavior has negative impact on urban economy by not only decreasing drivers' income but also generating additional vehicle miles traveled. Thus, repositioning available drivers to potential locations with near-future high demand, i.e., to balance supply and demand, becomes the key challenge faced by the taxi and for-hire market, including e-hailing platforms. 


\subsection{Problem statement}


This paper aims to solve the multi-driver repositioning task, 
in which a large number of idle drivers make sequential decisions on where to reposition themselves over a planning horizon, so that they can seek for the next passengers and maximize their individual cumulative rewards. 
When one idle driver has to make a sequential decision of repositioning while others are doing so simultaneously, 
competition among drivers arises.

Markov decision process (MDP) and reinforcement learning (RL) have become popular tools for sequential decision-making. 
Single-agent RL has been used in several studies \citep{gao_optimize_2018, okeeffe_using_2019}  to solve the single-driver repositioning problem. 
However, single-agent RL assumes each agent is an independent learner and ignores the impact of competition on each individual agent's policy.  
To capture competition among multiple agents in the process of repositioning, this paper employs a multi-agent reinforcement learning (MARL) approach. 


Nevertheless, this paper goes beyond simply modeling multi-driver repositioning with MARL. 
We notice that every driver plays a non-cooperative game, meaning each driver's goal is to select a sequence of reposition strategies to maximize her individual net profit, given other drivers select their own optimal reposition strategies. 
Accordingly, 
the direct application of MARL to the multi-driver system under a given reward function will likely yield a suboptimal equilibrium for the system, thanks to the selfishness of drivers. 
Therefore, the ultimate goal of this paper is to proposes a reward design scheme with which a more desired equilibrium can be reached. We would like to emphasize the difference between ``reward" and ``reward design".
In this paper, the reward for an idle driver is simply the net monetary return one driver earns, 
while the reward design scheme is imposed by a higher-level planner (e.g., the TNC platform or city planners) to modify drivers' monetary return.  
Analogous to the role of congestion pricing, reward design aims to spatially redistribute drivers, driven by individual reward maximization, 
through an external cost imposed to those who tend to move towards already oversupplied spots. 
Under the external cost, drivers still behave selfish but their movement is now governed by a modified reward, 
which consequently balances demand and supply 
and drives the system towards a more desirable equilibrium.



To justify that reward design can be an effective way of balancing supply and demand, 
we would like to briefly introduce our problem set-up below. 
We assume that each driver, behaving like an intelligent agent, selects a sequence of optimal reposition policies to maximize its own cumulative reward, which is the monetary income. 
Usually a TNC platform charges a commission fee to drivers. 
Thus, the revenue a TNC driver receives is an order's fare minus the commission fee. 
The TNC platform can adjust the percentage of fares going to drivers, in order to balance demand and supply. 
For instance, if the platform notices that excessive drivers try to move to one hotspot to compete for a limited number of orders, the platform can increase the commission fee in that grid to prevent oversupply. 
Drivers are still free to reposition themselves toward that grid, but adding a higher commission fee would reduce their long-run accumulative rewards, and accordingly, influence their reposition policies, which hopefully discourage some drivers from moving to that grid. 

\subsection{Literature review}


In recent years, we have witnessed a growing body of literature on TNC using reinforcement learning, including 
	pricing mechanism design \citep{pandey_multiagent_2018, pandey_deep_2019}, vehicle routing \citep{bazzan_multiagent_2016, nazari_reinforcement_2018, peng_deep_2020}, ride order dispatching \citep{li_efficient_2019, tang_deep_2019, ke_optimizing_2019, zhou_multi-agent_2019}, and driver repositioning \citep{gao_optimize_2018, lin_efficient_2018, yang_multiagent_2020}. This study focuses on the multi-driver repositioning task.  

The essence of the repositioning task is to provide recommendations to idle drivers on where to find the next passenger. 
Some recommender systems have been proposed for drivers \citep{ge_energy-efficient_2010,hwang_effective_2015,yuan_where_2011,qu_cost-effective_2014}. These studies extracted
useful aggregated statistical quantities such as taxi demand and travel time from historical data and recommended a next cruising location \citep{ge_energy-efficient_2010}, a sequence of
potential pickup points \citep{hwang_effective_2015}, a driving route \citep{qu_cost-effective_2014}, or a route and a location \citep{yuan_where_2011}. 
Although the aforementioned studies provide effective recommendations of the next cruising route or location to drivers at the immediate next step, they are nearsighted and fall short of capturing the future long-run payoffs. To capture the effect of future rewards on the recommendation at the immediate next step, various Markov decision process (MDP) based  approaches have been proposed to model idle drivers' passenger searching process \citep{rong_rich_2016,zhou_optimizing_2018,verma_augmenting_2017,gao_optimize_2018,yu_markov_2019,shou_optimal_2020}. 
In an MDP, a driver is the agent who makes decisions of where to go next and interacts with the environment. 
The agent aims to derive an optimal policy which maximizes her expected cumulative reward. When the environment is known to the agent, dynamic programming can be used to solve the MDP and derive an optimal policy. When the dynamic environment is unknown to the agent, 
reinforcement learning (RL) algorithms such as Q-learning and temporal difference learning \citep{sutton_introduction_1998} can be employed to derive an optimal policy.

The competition among multiple agents is, however, neglected in the aforementioned MDP models due to their single-agent setting, resulting in overly optimistic optimal policies. In other words, one agent cannot earn the full amount of the expected reward by following the policy derived in the single-agent setting. 
In a dynamic environment involving a group of agents, multiple agents interact with both the shared environment and other agents. Multi-agent reinforcement learning (MARL) \citep{busoniu_multi-agent_2010} thus fits naturally well in this multi-agent system (MAS). 
Recently, MARL has been attracting significant attention due to its success in tackling high dimensional and complicated tasks such as playing the game of Go \citep{silver_mastering_2016, silver_mastering_2017}, Poker \citep{brown_superhuman_2018, brown_superhuman_2019}, Dota 2 \citep{OpenAI_dota}, and StarCraft II \citep{vinyals_grandmaster_2019}. 

MARL tasks can be broadly grouped into three categories, namely, fully cooperative, fully competitive, and a mix of the two, depending on different applications \citep{zhang_multi-agent_2019}: (1) In the fully cooperative setting, agents collaborate with each other to optimize a common goal; (2) In the fully competitive setting, agents have competing goals, and the return of agents sums up to zero; (3) The mixed setting is more like a general-sum game where each agent cooperates with some agents while competes with others. For instance, in the video game \emph{Pong}, an agent is expected to be either fully competitive if its goal is to beat its opponent or fully cooperative if its goal is to keep the ball in the game as long as possible \citep{tampuu_multiagent_2017}. A progression from fully competitive to fully cooperative behavior of agents was also presented in \cite{tampuu_multiagent_2017} by simply adjusting the reward. 


A key challenge arises 
in MARL when independent agents have no knowledge of other agents, that is, the theoretical convergence guarantee is no longer applicable since the environment is no longer Markovian and stationary \citep{matignon_independent_2012, nguyen_deep_2018}. 
To tackle this issue, one way is to exchange some information among agents. 
In some contexts, agents actually exchange information with their peers through some coordination. 
For example, in the game of a team of hunters capturing a team of preys, 
\cite{tan_multi-agent_1993} proposed multiple ways to enable coordination among agents and 
concluded that the performance of the hunter agents can be better off through some coordination.  
However, in other contexts such as the driver repositioning system, agents only have access to their own information. Thus, information exchange among agents involves a central controller which collects the information of all agents and disseminates it to agents. Agents update their value functions and policies based on the provided information from the central controller and their local observations. This is the centralized learning (i.e., based on global information) and decentralized execution (i.e., based on local observation) paradigm, which has become increasingly popular in recent research \citep{foerster_learning_2016, lowe_multi-agent_2017, lin_efficient_2018, li_efficient_2019}. 

While training is stabilized conditioning on the information of other agents such as joint state and joint action in the centralized training paradigm, 
scalability becomes a critical issue in MARL because the joint state space and joint action space grow exponentially with the number of agents. To make MARL tractable when a large number of agents coexist, \cite{yang_mean_2018} employed the mean field theory to simplify the interaction among agents. 
The basic idea is, from the perspective of an agent, to treat other agents as a mean agent.   
Thus, the complexity of interactions among a large number of agents is substantially eased by reducing the dimension in the Q-value function. The large scale MARL with hundreds of or even thousands of agents becomes solvable. To investigate the large-scale order dispatching problem where thousands of agents are present, \cite{li_efficient_2019} adopted a mean field approximation and proposed to take the average response from neighboring agents as a proxy of the interaction between the agent and other agents. 

\subsection{Research gaps}


\begin{sidewaystable}
   \centering\caption{Existing research of order distpatching and driver repositioning using MARL}
   \label{tab:marl}
	\begin{tabular}{p{78pt} p{50pt}  p{35pt} p{130pt} p{110pt} p{80pt} p{30pt} p{80pt}} 
		\hline
		Research problem &  Reference & Agent & State representation & Reward & Algorithm & Reward design & Gap\\ \hline
		\multirow{4}{*}{Order dispatching} & \cite{li_efficient_2019} &  Driver & (grid\_id, time, on-trip flag) & Fare, destination potential, pickup distance & Mean field actor-critic & No &  - \\ 
		&  \cite{tang_deep_2019} & Driver & (grid\_id, time, dynamic features, static features) & Fare & Semi MDP and cerebellar value networks & No & - \\ 
		& \cite{ke_optimizing_2019} & Passenger & (grid\_id, cumulative waiting time, expected distance to matched driver, global distribution of supply and demand) & A constant reward for a match, cost & Spatial-temporal multi-agent actor-critic &  No & -  \\
		& \cite{zhou_multi-agent_2019} &  Driver & (grid\_id, number of idle vehicles, number of valid orders, distribution of orders' destinations) & Fare (proportional to distance) & Action selection Q-learning & No & - \\ \hline
		Driver repositioning and order dispatching & \cite{jin_coride_2019} &  Grid & (number of vehicles, number of orders, entropy, number of vehicles to reposition, distribution of order features) & Accumulated driver income, order response rate & Deep deterministic policy gradient & No & - \\
		\hline 
		\multirow{3}{*}{Driver repositioning } & \cite{lin_efficient_2018} &  Driver & (grid\_id, time, global state) & Fare (averaged in each grid), fuel cost & Contextual actor-critic & No & Using global information (i.e. global state)\\ 
		&  \cite{yang_multiagent_2020} & Driver  & (grid\_id) & 100 points if the chosen action leads to a balance between demand and supply otherwise -1 & WoLF(Win or Learn First) policy hill-climbing & No & Not optimizing cumulative reward \\
		&This work & Driver & (grid\_id, time) & Fare & Mean field actor-critic & Yes &  \\ \hline
 \hline
	\end{tabular}
\end{sidewaystable}

Recent studies have successfully applied MARL to order dispatching and driver repositioning problems. Table~\ref{tab:marl} summarizes these papers in terms of MARL set-up and algorithms. 
	Because this paper is focused on repositioning, we will mainly highlight the research gaps that exist in the studies on multi-driver repositioning. 

There are a few studies that have employed MARL for the task of driver repositioning \citep{lin_efficient_2018, yang_multiagent_2020}. 
For each time period, based on a difference matrix between the supply and the predicted demand, \cite{yang_multiagent_2020} adopted a WoLF (Win or Learn First) policy hill-climbing algorithm to reposition cooperative drivers with the goal of balancing demand and supply. Cumulative reward of drivers over multiple time periods, however, is not optimized. \cite{lin_efficient_2018} proposed a contextual actor-critic model to reposition thousands of drivers with each driver aims to maximize her cumulative reward. To make training tractable, a global state which includes the overall distribution of demand and supply is used. Global information, however, may not be obtainable from the perspective of an agent, especially during execution. 
To fill the above gaps, this paper aims to tackle the multi-driver repositioning task only using local observation of each driver.

Although the approaches listed in Table~\ref{tab:marl} are efficient under a given reward function, the reached equilibrium is very likely to be suboptimal from the overall perspective of the system. In other words, there may exist another reward function that yields a better equilibrium in terms of some metrics such as gross mechandise volume (GMV) and order response rate (ORR) in the context of taxi hailing.  One possible approach to achieve a better equilibrium is to intentionally craft a reward function that is aligned with the goal of the overall system. For example, \cite{li_efficient_2019} proposed the order destination potential as a component in the reward function to enable some cooperation among drivers and discourage drivers from being selfish. Human-drivers, however, are selfish in nature and focus on their own monetary return. Thus, the embeded goal of the system in the predefined reward function may not reflect the intrinsic interest of drivers, and therefore drivers may not follow the derived optimal policy. 
To address the above limitations of a predefined reward function, this study aims to achieve a more desirable equilibrium by adjusting drivers' monetary return through some realistic measures of the system. For example, e-hailing platforms can adjust the platform service charge (aka the commission fee) to achieve a better GMV or ORR. 
Likewise, city planners commonly use congestion pricing to drive the traffic system performance towards a system optimum in transportation network design problems \citep{yang1998models,zhang2004optimal,meng2012impact,di2014braess,di2016second,di2017ridesharing,di2018link}. 
We call these measures as ``\textbf{reward design}" mechanism.


In this paper, we show that by integrating a reward design mechanism which adjusts the monetary return that a driver earns, 
a desirable equilibrium can be reached in this intrinsically large-scale non-cooperative system. The desirable equilibrium refers to a Nash equilibrium where each independent and selfish agent's strategy is the best-response to other agents' strategies and will produce better overall performance of the system. 
\cite{mguni_coordinating_2019} proposed a two-layer architecture with an incentive designer as the upper layer and a potential game as the lower layer and formulated the incentive designer's problem as an optimization problem. In contrast, the MARL problem in our context may not be able to be transformed as a potential game, complicating computation of its equilibrium. 

\subsection{Contributions of this paper}

The major contributions of this paper are as follows: 
\begin{enumerate}
	\item To achieve a better equilibrium for the overall system, a reward design scheme is proposed, formulated as the upper level optimization in a bilevel mathematical program, to adjust drivers' monetary return for social optima. 

	\item With the repositioning task as the lower level in the bilevel mathematical program, a mean field actor-critic algorithm is employed to enable the learning involving thousands of agents. 
	A Bayesian optimization algorithm is then developed to solve the bi-level problem.
	\item In the case study of taxi driver repositioning under congestion pricing, a multiclass MARL is developed to capture the intrinsic behavioral difference between yellow taxis and green taxis. 
\end{enumerate}

The remainder of the paper is organized as follows.
Section~(\ref{sec:single}) introduces the single-agent actor-critic algorithm, which is a stepping stone for MARL. Section~(\ref{sec:mdmarl}) presents the mean field multi-agent reinforcement learning algorithm. Section~(\ref{sec:rd}) presents a reward design mechanism and formulates a bilevel optimization problem. Section~(\ref{sec:case}) presents the result and validates the effectiveness of the proposed reward design. Section~(\ref{sec:conclu}) concludes. 




\section{Single agent reinforcement learning} \label{sec:single}

As a stepping stone, we first introduce the single agent reinforcement learning where only one agent interacts with the environment. 

\subsection{Problem definition}
A Markov decision process (MDP) \citep{puterman_markov_1994}
is typically specified by a tuple $(S, A, R, P, \gamma)$, where $S$ denotes the state space, $A$ stands for the allowable actions, $R:S \times A \times S \rightarrow \mathbb{R}$ collects rewards, $P:S \times A \times S \rightarrow [0, 1]$ denotes a state transition probability from one state to another, and $\gamma \in [0,1]$ is a discount factor. A general MDP proceeds simply as follows. Starting from the initial state, the agent specifies an action $a\in A$ whenever the agent is in a state $s\in S$. The agent then transits into a new state $s'\in S$ with probability $P(s'|s,a)$ and observes an immediate reward $r(s,a,s')$ by obeying the dynamics of the environment. Then the process repeats until a terminal state is reached. A policy $\pi:S\times A \rightarrow [0,1]$ simply maps from state $s\in S$ to the probability of taking action $a\in A$ in state $s$, i.e., $\pi(a|s)$. The goal of solving an MDP is to derive an optimal policy $\pi^*$ so that the agent can maximize her long term expected reward by following the policy. 
In reinforcement learning problems, the transition probability matrix $P$ is commonly unknown, and the agent learns about $P$ from its interaction with the environment. 

Denote $V^{\pi}(s)$ as the state value, which is the expected cumulative reward that an agent can earn by starting from state $s$ and following a policy $\pi$. $V^{\pi}$ can be recursively given as \citep{sutton_introduction_1998}
$V^{\pi}(s) = \mathbb{E}_{a \sim \pi(\cdot|s), s' \sim P(\cdot|s,a)} \left[ r(s,a,s') + \gamma V^{\pi}(s') \right]$.
Denote $Q^{\pi}(s,a)$ as the state-action value, which is the expected cumulative reward that an agent can earn by starting from state $s$, taking action $a$, and following a policy $\pi$. $Q^{\pi}$ is related with $V^{\pi}$ through
$Q^{\pi}(s,a) = \mathbb{E}_{s' \sim P(\cdot|s,a)} \left[r(s,a,s') + \gamma V^{\pi}(s') \right]$. 

The optimal value $V$ can then be written as
$V(s) = max_{\pi}V^{\pi}(s), \forall s \in S$. 
The Bellman optimality equation is given as \citep{sutton_introduction_1998}: 
\begin{equation*}
V(s) =  max_a \mathbb{E}_{s' \sim P(\cdot|s,a)} [r(s,a,s') + \gamma V(s')], 
\end{equation*}
where the optimal state-action value is
$Q(s,a) = \mathbb{E}_{s' \sim P(\cdot|s,a)} [r(s,a,s') + \gamma V(s')]$. 

Our task is then to derive an optimal policy $\pi^*$ (i.e., to solve the MDP) with which the agent can optimize its expected cumulative reward. 

To demonstrate how to formulate an MDP under the context of e-hailing driver reposition, we will use examples on a 2-by-2 grid world throughout the paper every time when models are introduced. 

\begin{exmp}
	(Single-Agent $2 \times 2$).
	The single-agent driver reposition is presented in Figure~(\ref{subfig:toy_example1}). 
	We adopt a grid world setup where the index of each grid (denoted as $l$) is shown at the upper left corner. The taxi icon denotes the driver, and the person icon is the passenger request with the corresponding fare shown above. The time beneath the driver and the passenger request records the current time of the driver and the appearance time of the passenger request, respectively. The dashed line with arrow shows the origin and destination of the passenger request.  In addition to the emergence of passenger requests (i.e., orders), the stochastic environment also has an order dispatching component.  Considering that the scope of this study is driver repositioning instead of order dispatching, we assume passenger requests are randomly assigned to available drivers within the same grid. 
	
	
	\begin{figure}[H]
		\centering
		\includegraphics[width=0.35\linewidth,height=0.25\textheight,keepaspectratio]{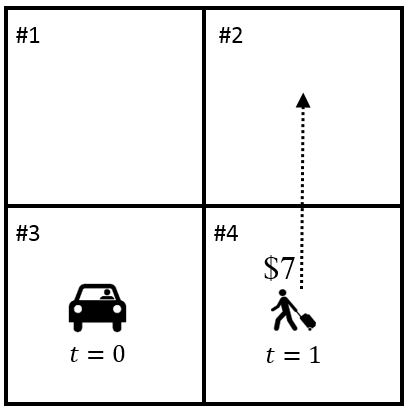}
		\centering 
		\caption{An illustrative example (Single agent)}
		\label{subfig:toy_example1}
	\end{figure}


	Note that there are a few studies solving vehicle routing problems based on a physical roadway network  \citep{liu_understanding_2013, yu_markov_2019}. Considering that the research scope of this paper is where to reposition drivers, we adopt an abstract network (i.e., a grid world setup). In other words, this paper is more focused on the direction along which a driver positions herself after dropping of previous passengers and not on the specific route the driver takes. 
	Using a real network representation for driver-respositioning will be left for future research.

	\emph{S.} The state of the driver consists of two components, namely, the grid index $l$ and current time $t$, i.e., $s = (l, t)$. For instance, the current state of the driver is $s = (\#3, 0)$ in this example. 
	
	\emph{A.} The allowable action of the driver is either moving into one of the neighboring grids or staying within the current grid. To be concise, we use the index of grid where the driver chooses to enter as the action. Suppose the driver decides to go rightward in the example, then we can denote $a = \#4$. We further assume it takes the driver one time step to enter grid $\#4$. In other words, the current time of the driver is $t = 1$ when the driver arrives in grid $\#4$.  
	
	
	
	\begin{note}
			At each time step, a taxi driver can only pick one of the surrounding grids or the current grid as her preferred spot to find a passenger. In other words, taxi drivers are only allowed local search at each time step. Although it seems to be restrictive, it is actually consistent with what real drivers do, and the reason is as follows. From the perspective of a real driver, she can pick a faraway grid as her destination. However, she can not jump directly from her current grid to that grid. Instead, the driver needs to specify a route (i.e., a sequence of grids) and repositions herself one grid each time along the chosen route.  
			If the driver meets a passenger before reaching her chosen destination, she has to take the passenger because she is not allowed to refuse a fare by law. From the perspective of an agent who is conducting local search, although she can not choose a faraway grid, she may end up searching in a faraway grid by repositioning herself one grid at a time. Actually, this definition of action is widely used in driver repositioning problems in the literature \citep{rong_rich_2016, lin_efficient_2018, shou_optimal_2020}.
	\end{note}

	\emph{P.} Considering the driver arrives in grid $\#4$ at time $t = 1$, and at the same time a passenger request appears in grid $\#4$ with $80\%$ probability. 
	If this driver is matched to the passenger and picks up the passenger, the driver will transit to the passenger's destination, which is grid $\#2$. 
	Denote the transition time from grid $\#4$ to grid $\#2$ as $\Delta t_{\#4 \rightarrow \#2}$. 
	We can define the new state $s' = (\#2, 1 + \Delta t_{\#4 \rightarrow \#2})$. 
	Then the transition probability from the state $s$ at time $0$ to the state $s'$ at time $1 + \Delta t_{\#4 \rightarrow \#2}$ is $80\%$, mathematically, 
	$P(s'|s, a) = 80\%$. 
	If there is no passenger request in grid $\#4$ at time $t = 1$, then the driver ends up in state $s' = (\#4, 1)$. 
	The transition probably becomes $P(s'|s, a) = 20\%$. 
	
	\begin{note}
	In the temporal component of state $s'$, we assume searching time is $1$, i.e., the time that one idle driver spent on cruising from her current grid to one of neighbor grids, and use $\Delta t_{\#4\rightarrow \#2}$ to denote the time that one on-duty driver spent on transporting the passenger to the destination. 
	Actually, 
	the travel time from passenger's origin to the destination is extracted from some historical data. In addition, the length of each time slice is $1$ in this example. 
	\end{note}
	
	\emph{R.} If we take the fare of the fulfilled passenger request as the reward, $r(s,a,s') = \$7$ in the example. 
	Based on the received reward at this step and the future cumulative reward, the driver chooses an action in the new state $s'$, and the state transition process repeats until a terminal state (i.e., $t = T$ where $T$ is a predefined ending time, say, the end of the driver's work time) is reached. 
	\qed
\end{exmp}

\subsection{Actor-Critic method}
To solve optimal policies, there are two types of methods, namely, value based or critic-only method and policy based or actor-only method. Value based and policy based methods are commonly used terminologies, but from now on we will use critic-only and actor-only methods for the purpose of introducing the actor-critic method. 

Critic-only methods aim to output the optimal policy $\pi$ through optimizing the state-action $Q(s,a)$ or the state value $V(s)$. 
Actor-only methods directly output an optimal policy $\pi$ without resorting to stored value functions $Q(s,a)$ or $V(s)$ as an intermediary. 
Both methods have pros and cons. 
Critic-only methods enjoy a low variance in the estimate of the state-action value but may lack guarantees on the optimality or near-optimality of the resulting policy if an optimal policy cannot be easily solved from value functions.  
Actor-only methods work well on continuous and large action spaces but may suffer from high fluctuation in policies \citep{konda_actor-critic_2003, grondman_survey_2012}. 
To overcome the shortcomings of these methods, actor-critic methods are developed to combine strengths of both methods 
\citep{konda_actor-critic_2003}. 

\begin{figure}[H]
	\centering
	\includegraphics[width=0.5\linewidth,height=0.25\textheight,keepaspectratio]{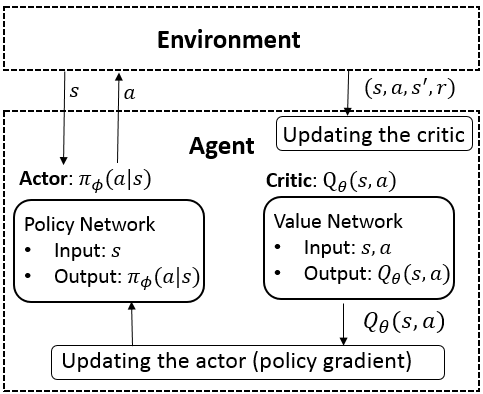}
	\centering 
	\caption{Actor-critic algorithm}
	\label{fig:ac}
\end{figure}

Figure~(\ref{fig:ac}) presents the architecture of the actor-critic algorithm. One agent, who has an actor and a critic, interacts with the environment. The agent observes its state $s$ from the environment and inputs $s$ to the actor that outputs the policy, i.e., a probability distribution over all possible actions. The agent samples an action $a$ from the probability distribution and takes action $a$ in the environment. Then the agent observes a state transition $s \rightarrow s'$ and receives a reward $r$ from the environment. Based on the one-step transition $s \rightarrow s'$ as well as action $a$ and reward $r$, the agent updates its critic. With the updated Q-value $Q_{\theta}(s,a)$, the agent updates its actor using policy gradient. Now we detail both the critic and the actor, respectively.


\noindent\textbf{Critic.} The critic takes as input state $s$ and action $a$ and outputs Q-value $Q(s,a)$. Q-learning is the most commonly used algorithm to update the Q value based on the state transition $s \rightarrow s'$ with reward $r(s,a,s')$ and updates the Q-value by
\begin{equation}
Q(s,a) \leftarrow Q(s,a) + \eta [r(s,a,s') + \gamma max_{a'}Q(s', a') - Q(s,a)]
\label{eqn:qupdate}
\end{equation}
where $\eta$ is the learning rate and $0 < \eta \leq 1$. If $\eta$ reduces over time properly, the Q-learning update converges \citep{sutton_introduction_1998}. Equation~(\ref{eqn:qupdate}), however, is only applicable to a finite and discrete state and action space. In other words, one needs to maintain a Q table with all possible combinations of $s$ and $a$, which is not tractable for a continuous and large state and action space. 
Therefore we need functional approximation to the original Q-value. 
Deep neural network, i.e., deep Q network (DQN), is one of the most popular value approximator \citep{mnih_human-level_2015}. 
Denote a deep neural network parameterized by $\theta$ as $Q_{\theta}(s,a)$, to approximate $Q(s,a)$. 
DQN updates its parameter $\theta$ by minimizing the loss 
\begin{equation}
\mathcal{L}(\theta) = \mathbb{E}_{s, a, s'} [(\underbrace{r(s,a,s') + \gamma max_{a'}Q_{\theta^-}(s',a')}_{\text{target}} - Q_{\theta}(s,a))^2]. \label{eqn:vloss}
\end{equation}
This problem can be solved by the gradient descent method, whose gradient is straightforward to compute as follows: $\nabla_{\theta} \mathcal{L(\theta)} = \mathbb{E}_{s,a,s'}[-\nabla_{\theta} Q_{\theta}(s,a) \times (r(s,a,s') + \gamma max_{a'}Q_{\theta^-}(s',a') - Q_{\theta}(s,a))]$, where the gradient is not taken with respect to the target. $Q_{\theta^-}$ is a target network which is a copy of $Q_{\theta}$. The target network $Q_{\theta^-}$ is not updated by the gradient descent and is copied from $Q_{\theta}$ after a certain number of steps to ensure training stability \citep{mnih_human-level_2015}. 
 
 
\noindent\textbf{Actor.} 
The actor takes as input state $s$ and outputs a probability distribution on all allowable actions in this state. 
Similar to how we use a value network to approximate Q-value, we can also use a deep neural network, i.e., policy network, to approximate the policy $\pi$. 
Denote the policy network parameterized by $\phi$ as $\pi_{\phi}(a|s)$. 
The goal of the actor is to maximize its expected cumulative reward, denoted as $\rho(\pi_{\phi}) = \sum_{t = 0}^T \gamma^t r^t$, where $r^t$ is the reward the actor receives at time $t$. 
To solve the optimal policy of the actor requires us to know its gradient. 
The gradient of the policy is complicated to solve and is given as \citep{sutton_policy_1999}
\begin{equation}
\nabla_{\phi}\rho(\pi_{\phi}) = \mathbb{E}_{s, a} [(\underbrace{Q_{\theta^-}(s,a) - b_{\theta^-}(s)}_{\text{advantage}}) \nabla_{\phi}log(\pi_{\phi}(a|s))], \label{eqn:pg}
\end{equation}
where 
$b_{\theta^-}$ is some baseline (e.g., $b_{\theta^-} = V_{\theta^-}$, i.e., the value function following the policy $V_{\theta^-}$), and $Q_{\theta^-}(s,a) - b_{\theta^-}(s)$ is called the advantage of a taken action $a$, a measure of the goodness of an action. 
If it is greater than zero, it means this taken action is generally good, otherwise it may be bad. Naturally, the underlying rationale in computing the policy gradient defined in Equation~(\ref{eqn:pg}) is to update the policy distribution to concentrate on potentially good action(s).
When the chosen action $a$ leads to a positive advantage, i.e., $Q_{\theta^-}(s,a) - b_{\theta^-}(s) > 0$, the policy is updated towards the direction of favoring action $a$. When the advantage is negative for action $a$, the policy is updated in the direction of against action $a$. 
%

To summarize, in addition to the policy network $\pi_{\phi}$, the actor-critic algorithm also maintains a value network $Q_{\theta}$ so that the calculation of the gradient of the policy in Equation~(\ref{eqn:pg}) directly uses the Q-function approximator $Q_{\theta}$, to ensure stability of policy update. The actor-critic algorithm simultaneously updates critic (by minimizing the loss given in Equation~(\ref{eqn:vloss})) and the actor (by the gradient given in Equation~(\ref{eqn:pg})) as more samples are fed in.

\section{Multi-agent reinforcement learning} \label{sec:mdmarl}


To tackle a real-world problem with multiple agents, the aforementioned single agent reinforcement learning falls short of capturing the coupling effects or the competition among multiple agents. In this section, we introduce a multi-agent reinforcement learning approach to model the multi-driver repositioning task. Drivers are assumed to be intelligent and perfectly rational, meaning they select the best repositioning strategies to maximize their cumulative rewards. Bounded rationality \citep{di_boundedly_2013, di_boundedly_2016} is left for future research. 

\subsection{Problem definition}

The multi-agent problem is modeled as a partially observable Markov decision process (POMDP) \citep{littman_markov_1994}, defined by a tuple $(S, O_1, O_2, \cdots, O_N, A_1, A_2, \cdots, A_N, P, R_1, R_2, \cdots, R_N, N, \gamma)$, where $N$ is the number of agents and $S$ is the environment state space. Environment state $\mathbf{s} \in S$ is not fully observable. Instead, agent $i$ draws a private observation $o_i \in O_i$ which is correlated with $\mathbf{s}$. $O_i$ is the observation space of agent $i$, yielding a joint observation space $O = O_1 \times O_2, \times \cdots \times O_N$, $A_i$ is the action space of agent $i \in \{1,2,\cdots, N\}$, yielding a joint action space $A = A_1 \times A_2 \times \cdots \times A_N$, $P:S\times A \times S \rightarrow [0, 1]$ is the state transition probability, $R_i:S\times A \times S \rightarrow \mathbb{R}$ is the reward function for agent $i$, and $\gamma$ is the discount factor. 

Agent $i \in \{1,2,\cdots,N\}$ uses a policy $\pi_i:O_i \times A_i \rightarrow [0,1]$ to choose actions after drawing observation $o_i$. 
After all agents taking actions, the joint action $\mathbf{a}$ triggers a state transition $\mathbf{s} \rightarrow \mathbf{s}'$ based on the state transition probability $P(\mathbf{s}'|\mathbf{s}, \mathbf{a})$. Agent $i$ draws a private observation $o_i'$ corresponding to $\mathbf{s}'$ and receives a reward $r_i(\mathbf{s}, \mathbf{a}, \mathbf{s}')$.  Agent $i$ aims to maximize its discounted expected cumulative reward by deriving an optimal policy $\pi^*_i$ which is the best response to other agents' policies. This process repeats until agents reach their own terminal state.

Due to the existence of other agents, the Q-value function for agent $i$ 
, i.e., $Q_i$, is now dependent on the environment state $\mathbf{s} \in S$ and the joint action $\mathbf{a} \in A$ of all agents
, i.e, 
\begin{equation}
Q_i = Q_i(\mathbf{s}, \mathbf{a}). \label{eqn:q}
\end{equation}
Similarly, the value function of agent $i$, i.e., $V_i = V_i(\mathbf{s})$, 
is dependent on the environment state $\mathbf{s}$. 

Subsequently, we will demonstrate how to formulate the multi-driver repositioning problem in MARL, building on the single-agent example developed in the previous section.

\begin{exmp}
	
(Multi-Agent $2 \times 2$). The multi-agent driver reposition is presented in Figure~(\ref{subfig:toy_example}). Same as before, a grid world setup is adopted. Now we have two drivers with their indices shown above the taxi icon and two passenger requests with fare presented above the passenger icon. The time beneath drivers and passenger requests records the current time of the driver and the appearance time of the passenger request, respectively. The dashed line with arrow shows the origin and destination of the passenger request.  

\begin{figure}[H]
	\centering
	\includegraphics[width=0.35\linewidth,height=0.25\textheight,keepaspectratio]{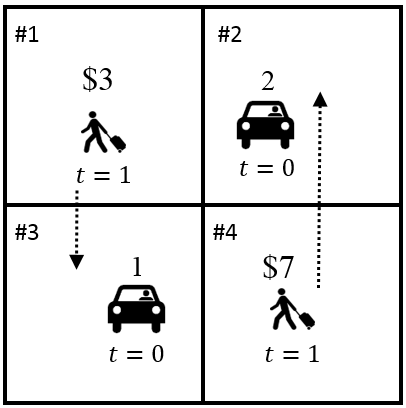}
	\centering 
	\caption{An illustrative example (Multi agent)}
	\label{subfig:toy_example}
\end{figure}

\emph{N.} There are $N = 2$ drivers moving around in the environment. We denote drivers by $\{1,2\}$. 

\emph{S.} The environmental state consists state information of both drivers. For driver $i$, her state $s_i$ is composed of her current location $l_i$ (i.e., the grid index based on a grid world setup) and current time $t$, i.e., $s_i = (l_i, t)$. The joint state of both drivers, i.e., the environment state $\mathbf{s} \in S$, at time $t$ is denoted as $\mathbf{s} = (s_1, s_2)$. In this example, at current time $t = 0$, $\mathbf{s} = ((\#3, 0), (\#2, 0))$.

\emph{A.} For driver $i$, her action $a_i \in A_i$ can be any of the five possible actions, i.e., moving into any of her four neighboring grids or staying in the current grid. The same as before, we use the index of grid where the driver chooses to enter as the action. 
The joint action of both drivers is $\mathbf{a} = (a_1, a_2)$. 
Assuming driver $1$ decides to go rightward (i.e, to enter grid $\#4$) and driver $2$ chooses to go leftward (i.e., to enter grid $\#1$), the joint action is $\mathbf{a} = (\#4, \#1)$. We further assume it then takes driver $1$ one time step to enter grid $\#4$ and driver $2$ one time step to enter grid $\#1$. In other words, after driver $1$ arrives in grid $\#4$ and driver $2$ arrives in grid $\#1$, the clock ticks one step forward and the current time is now $t = 1$.

\emph{P.} The joint action $\mathbf{a}$ triggers a state transition $\mathbf{s} \rightarrow \mathbf{s}'$ with some probability according to the state transition function, i.e., $P(\mathbf{s}|\mathbf{s}',\mathbf{a})$. Driver $1$ gets matched to the passenger request in grid $\#4$ at $t = 1$, loads up the passenger, and drives to the destination of the passenger. Driver $1$ then arrives in a new state $s_1' = (\#2, 1 + \Delta t_{\#4 \rightarrow \#2})$ where $\Delta t_{\#4 \rightarrow \#2}$ is the transition time from grid $\#4$ to grid $\#2$. Similarly, driver $2$ gets matched to the passenger request in grid $\#1$ at $t = 1$, loads up the passenger, and drives to the destination of the passenger. Driver $2$ then arrives in a new state $s_2 = (\#3, 1 + \Delta t_{\#1 \rightarrow \#3})$ where $\Delta t_{\#1 \rightarrow \#3}$ is the transition time from grid $\#1$ to grid $\#3$. $\mathbf{s}' = ((\#2, 1 + \Delta t_{\#4 \rightarrow \#2}), (\#3, 1 + \Delta t_{\#1 \rightarrow \#3}))$. In this simple example, $P(\mathbf{s'}|\mathbf{s}, \mathbf{a}) = 1$ due to the deterministic appearance of passenger requests.

\emph{R.} Along with the state transition, each driver receives a reward, i.e., $r_i$. The reward function $r_i$ for each agent $i \in \{1,2\}$ is simply the fare of the fulfilled passenger request, i.e., $r_1 = \$7$ and $r_2 = \$3$.  
\qed 
\end{exmp}

This example will be revisited later in this section to illustrate the algorithm. 

\subsection{Techniques to simplify the Q-value function}

The dependency of the Q-value of an agent $i$ on other agents' states and actions, as shown in Equation~(\ref{eqn:q}), however, introduces prohibitively high difficulties in learning the optimal Q-value. The main reasons are two-fold. First, although each agent draws its private observation $o_i$ from the environment state $\textbf{s}$, $\textbf{s}$ cannot be observed by any agent, i.e., $\textbf{s}$ is unknown. 
Second, one agent does not observe the actual actions taken by all agents, i.e., $\mathbf{a}$ is unknown. 

To make the Q-value of an agent in the multi-agent system tractable, the dependency of the Q-value on the environment state $\mathbf{s}$ and joint action $\mathbf{a}$ needs to be simplified.  
A very natural approach, inspired by the single-agent setting, is independent learning where 
each agent $i$ only has information about its own observation $o_i$ and action $a_i$ but has no information about other agents. 
Thus, the Q-value function of agent $i$ is reduced to 
\begin{equation}
Q_i = Q_i(o_i, a_i).
\end{equation}
In other words, private observations and joint action of other agents are not used by agent $i$.
After all agents choosing actions, the joint action $\mathbf{a}$ triggers a state transition. Agent $i$ then draws a new private observation $o_i'$ and receives a reward $r_i$. 

The independent learning algorithm, although is intuitive and simple, can be unstable and hard to reach convergence since the environment is no longer Markovian and stationary due to the appearance of other agents \citep{matignon_independent_2012}. 

\subsubsection{Centralized training and decentralized execution}

To make the training more stable and ensure convergence, we employ the centralized training and decentralized execution paradigm \citep{foerster_learning_2016, lowe_multi-agent_2017, lin_efficient_2018, li_efficient_2019}. 
In this paradigm, to train the policy of agents, we assume these agents know the global information such as the joint observation and/or joint action. 
In other words, in addition to observation $o_i$ and action $a_i$, agent $i$ also has access to the observations and/or actions of other agents during training. 
While in the execution phase
, decentralized testing or execution is implemented, meaning they would not have access to the global information anymore. 
To realize this paradigm, the aforementioned actor-critic algorithm naturally fits in, 
because we can apply global information to the critic, i.e., joint observation and joint action in $Q_i$, in the training phase, 
while feeding local information to the actor, i.e., $o_i$ in $\pi_i$, in the execution phase. 
Decentralized execution becomes possible because only actors are used in execution.

Then the Q-value function of agent $i$ becomes
\begin{equation}
Q_i = Q_i(o_i, o_{-i}, a_i, a_{-i}),
\end{equation}
where $o_{-i} = (o_1, \cdots, o_{i - 1}, o_{i + 1}, \cdots, o_N)$ and $a_{-i} = (a_1, \cdots, a_{i - 1}, a_{i + 1}, \cdots, a_N)$ denote the joint observation and joint action of all agents except agent $i$, respectively. 

In the context of e-hailing driver repositioning,
considering the definition of the action, which is the index of the grid where the driver chooses to enter, the Q-value function of driver $i$, i.e., $Q_{i}$, does not depend on the joint observation of other drivers, i.e., $o_{-i}$. 
Explanations are as follows. When driver $i$ chooses action $a_i = l$ based on its observation $o_i$, driver $i$ then enters grid $l$. At the same time, other drivers also enter some grid based on their joint action $a_{-i}$ regardless of their joint observation $o_{-i}$. The Q-value function of driver $i$ only depends on the current distribution of drivers, which has been determined by their joint action $a_{-i}$. Therefore it is the joint action $a_{-i}$ which affects $Q_i$. The Q-value function is thus further reduced to
\begin{equation}
Q_i = Q_i(o_i, a_i, a_{-i}).
\end{equation}


\subsubsection{Mean field approximation}

The centralized training and decentralized execution paradigm, however, can easily become intractable due to the exponential increase in the joint action space with the increasing number of agents. For example, the size of the joint action space easily blows up for $N$ agents with $|A|$ possible actions (i.e., $|A|^{N}$ possibilities).
To simplify the interaction among agents, we adopt the mean field approximation. The basic idea of the mean field approximation is to simplify the complicated interaction between one agent and all other agents by a pairwise interaction between the agent and a virtual mean agent which is formed by the neighboring agents of the agent. 
Thus, the complexity of interactions among a large number of agents is substantially eased by reducing the dimension in the input of the Q-value function. Therefore the large scale MARL with hundreds of or even thousands of agents becomes solvable.

To be more precise, we provide brief explanations that lead to the applicability of the mean field approximation in MARL as described in \cite{yang_mean_2018}. First, from the perspective of agent $i$, the multi-agent effect or competition effect mainly comes from its neighboring agents, i.e., $Q_i(o_i, a_i, a_{-i}) \approx \dfrac{1}{N_i} \sum_{k \in N(i)} Q_i(o_i, a_i, a_k)$, where $N(i)$ denotes the neighboring agents of agent $i$. 
However, it is still cumbersome to compute $Q_k, k \in N(i)$ for the neighboring agents of agent $i$ if this number is large.  
Define a mean action $\bar{a}_i$, which is a proxy of the actions taken by the neighboring agents. 
Accordingly, $Q_i(o_i, a_i, a_{-i})$ can be further simplified to $Q_i(o_i, a_i, \bar{a}_i)$ when Taylor expansion is applied, which is 
\begin{equation}
Q_i \approx \dfrac{1}{N_i} \sum_{k \in N(i)} Q_i(o_i, a_i, a_k) \approx Q_i(o_i, a_i, \bar{a}_i).
\end{equation} 
Interested readers can refer to \cite{yang_mean_2018} for a detailed explanation and proof. 

\begin{exmp} (Multi-Agent $2 \times 2$). The mean action $\bar{a}_i$ of the neighboring drivers of driver i is defined as the demand to supply ratio in the grid where driver $i$ is entering. 	Assuming both drivers choose action $\#4$, i.e., $a_1 = a_2 = \#4$ in the multi-agent $2 \times 2$ example shown in Figure~(\ref{subfig:toy_example}), there are 2 drivers and 1 passenger request in grid $\#4$ after both drivers enter grid $\#4$. The mean action for both drivers is thus $\bar{a}_1 = \bar{a}_2 = \frac{1}{2} = 0.5$. 
This definition of mean action captures the level of competition in a grid. A larger mean action $\bar{a}_i$ denotes a higher demand to supply ratio and lower level of competition, and vice versa. 
	\qed
\end{exmp}

\subsection{Mean field actor-critic algorithm}

As previously mentioned, each agent $i \in \{1, 2, \cdots, N\}$ maintains a policy network $\pi_i$ (i.e., the actor) and a Q-value network $Q_i$ (i.e., the critic). 
For a real-world multi-agent task, however, there are typically hundreds of or even thousands of agents, resulting in a prohibitively large number of deep neural networks to maintain, which is not computationally tractable. 
To address this issue, we assume drivers are homogeneous and they share the same state space, action space, and reward function. 
We believe this assumption is reasonable when agents are anonymous and one's influence to the entire system performance is negligible. 
While heterogeneity across agents does exist when agents have different reward functions \citep{shou_optimal_2020}, this study aims to reveal some insights of multi-driver repositioning by learning a policy for the generic taxi population. A more personalized reposition scheme that captures the heterogeneity across the driver population is left for future research. 

 The multi-agent task can then be largely simplified by sharing both the actor and the critic among drivers, i.e., $Q_1 = Q_2 = \cdots = Q_N = Q$ and $\pi_1 = \pi_2 = \cdots \pi_N = \pi$. 

After adopting the mean field approximation
, the loss function for the critic, which was presented in Equation~(\ref{eqn:vloss}) for the single-agent setting, now becomes
\begin{equation}
\mathcal{L}(\theta) = \mathbb{E}_{o_i, a_i, o_i'} (r(o_i,a_i,o_i') + \gamma max_{a_i'} \mathbb{E}_{\bar{a}_i'}[Q_{\theta^-}(o_i',a_i',\bar{a}_i')] - Q_{\theta}(o_i,a_i,\bar{a}_i))^2. \label{eqn:vloss_mean}
\end{equation}
The only difference is the incorporation of the mean action $\bar{a}$ into the Q-value function approximation. Similarly, the gradient of the policy, which was presented in Equation~(\ref{eqn:pg}) for single-agent setting, is now
\begin{equation}
\nabla_{\phi}\rho(\pi_{\phi}) = \mathbb{E}_{o_i, a_i} [(\mathbb{E}_{\bar{a}}[Q_{\theta^-}(o_i,a_i,\bar{a}_i)] - V(o_i)) \nabla_{\phi}log(\pi_{\phi}(a_i|o_i))], \label{eqn:pg_mean}
\end{equation}
where the baseline $V(o_i) = \mathbb{E}_{a_i, \bar{a}_i} [Q_{\theta^-}(o_i, a_i, \bar{a}_i)]$.




\begin{algorithm}[H]
	\caption{Mean field actor-critic algorithm}
	\label{alg:qlearning}
	\begin{algorithmic}[1]
		\State Input: exploration parameter $\epsilon = \epsilon_0$, 
		target network update period $\tau$, learning rate $\eta=\eta_0$, and a preset number of samples $K$
		\State Initialize a deep neural network $Q_{\theta}(o,a,\bar{a})$, parameterized by $\theta$, for the critic and a deep neural network $\pi_{\phi}(a|o)$, parameterized by $\phi$, for the actor
		\State Initialize a target Q-value network $Q_{\theta^-}$ with parameter $\theta^-=\theta$ and a target policy network $\pi_{\phi^-}$ with $\phi^-=\phi$
		\State Initialize a replay buffer $B $
		\State Set $I=0$
		\Repeat
		\State Randomly initialize a starting grid for all agents, and each agent $i \in \{1,2,\cdots,N\}$ draws a private observation $o_i$
		\State Set $t = 0$
		\Repeat
		\State Sample a value $x$ from a uniform distribution which is defined on $[0,1]$
		\If{$x < \epsilon$}
		\State Select an action $a_i$ from the allowable action space randomly for all available agents
		\Else
		\State Select an action $a_i$ greedily according to the policy $\pi_{\phi^-}$ for all available agents
		\EndIf
		\State Each available agent takes its action $a_i$, observes a reward $r_i$ and a mean action $\bar{a}_i$, and draws a new observation $o_i'$
		\State Store the experience tuple $\{o_i, a_i, r_i, \bar{a}_i, o'_i\}$ into the replay buffer $B$
		\State $t \leftarrow t + 1$
		\Until{$t = T$}
		\State Sample $K$ sample experience tuples from the buffer $B$
		\State Update the critic $Q$ by minimizing the loss defined Equation~(\ref{eqn:vloss_mean}) 
		\State Update the actor $\pi$ using the gradient defined in Equation~(\ref{eqn:pg_mean})
		\State Decrease the exploration parameter $\epsilon$ 
		\State Decrease the learning rate $\eta$
		\State $I = I + 1$
		\If{$I~\text{mod}~\tau = 0$}
		\State Update the parameter of target networks, i.e., $\theta^- \leftarrow \theta$ and $\phi^- \leftarrow \phi$
		\EndIf
		\Until{the algorithm converges}
		\State Return the optimal policy $\pi$
	\end{algorithmic}
\end{algorithm}

The mean field actor-critic algorithm is presented in Algorithm~\ref{alg:qlearning}. With the input of some parameters such as the exploration parameter $\epsilon$, target network update period $\tau$, and learning rate $\eta$, and a preset number of samples $K$, a neural network $Q_{\theta}$ for the critic and a neural network $\pi_{\phi}$ for the actor are initialized. Target networks $Q_{\theta^-}$ and $\pi_{\phi^-}$ are copied from $Q_{\theta}$ and $\pi_{\phi}$, respectively. An experience replay buffer $B$ is also initialized to store experience tuples of all agents. Now we let agents repetitively interact with the environment and update the actor and the critic as follows. Agents are randomly placed in the environment initially. All agents choose actions according to the $\epsilon$-greedy method. That is, agents have $\epsilon$ probability of choosing actions randomly and $1 - \epsilon$ probability of choosing actions according to the target policy network $\pi_{\phi^-}$. Each agent $i$ then executes the chosen action $a_i$ in the environment, observes a reward $r_i$ and a mean action $\bar{a}_i$, and draws a new observation $o_i'$. This experience tuple $\{o_i, a_i, r_i, \bar{a}_i, o_i'\}$ is stored in the replay buffer $B$. All idle agents then choose actions again and repeat the aforementioned procedure until reaching the terminal state with $t = T$. After collecting experience tuples of all agents, $K$ sample experience tuples are randomly sampled from the replay buffer $B$ and used to train neural networks $Q_{\theta}$ and $\pi_{\phi}$ by Equations (\ref{eqn:vloss_mean}) and (\ref{eqn:pg_mean}).



\begin{exmp} (Multi-Agent $2 \times 2$). Now we apply the mean field actor-critic algorithm to the multi-driver example shown in Figure~(\ref{subfig:toy_example}). Figure~(\ref{fig:ac_mean}) presents the architecture of the mean field actor-critic algorithm particularly for the context of multi-driver repositioning. 
Homogeneous agents, who share a common actor and a common critic, interact with the environment. The shared actor is a multilayer perceptron with 32 neurons in its hidden layer and takes as input observation $o_i$ and outputs a five dimensional vector denoting the probability distribution of taking five actions. Similarly, the shared critic takes as input $(o_i, a_i, \bar{a}_i)$ and outputs the Q-value. 
During training, agent $i \in \{1,2,\cdots,N\}$ draws its private observation $o_i$ from the environment and inputs $o_i$ to the actor which outputs a probability distribution over actions. Agent $i$ samples an action $a_i$ from the probability distribution and takes the sampled action in the environment. Joint action of all agents $\mathbf{a}$ triggers a state transition $\mathbf{s} \rightarrow \mathbf{s}'$ in the environment. Agent $i$ then observes the mean action $\bar{a}_i$, draws a new observation $o_i'$, and receives a reward $r_i$ from the environment. The agent then uses $(o_i, a_i, o_i', r_i, \bar{a}_i)$ to update the shared critic by minimizing the loss presented in Equation~(\ref{eqn:vloss_mean}). Based on the advantage calculated from the critic, agent $i$ updates the shared actor using the gradient presented in Equation~(\ref{eqn:pg_mean}). 

The aforementioned training process is centralized because the mean action used in the critic is actually some global information. During execution, agents only need to use the updated actor, which only takes as input the local information, i.e., the private observation. In other words, the shared critic is not used in execution.  

\begin{figure}[H]
	\centering
	\includegraphics[width=0.99\linewidth,height=0.5\textheight,keepaspectratio]{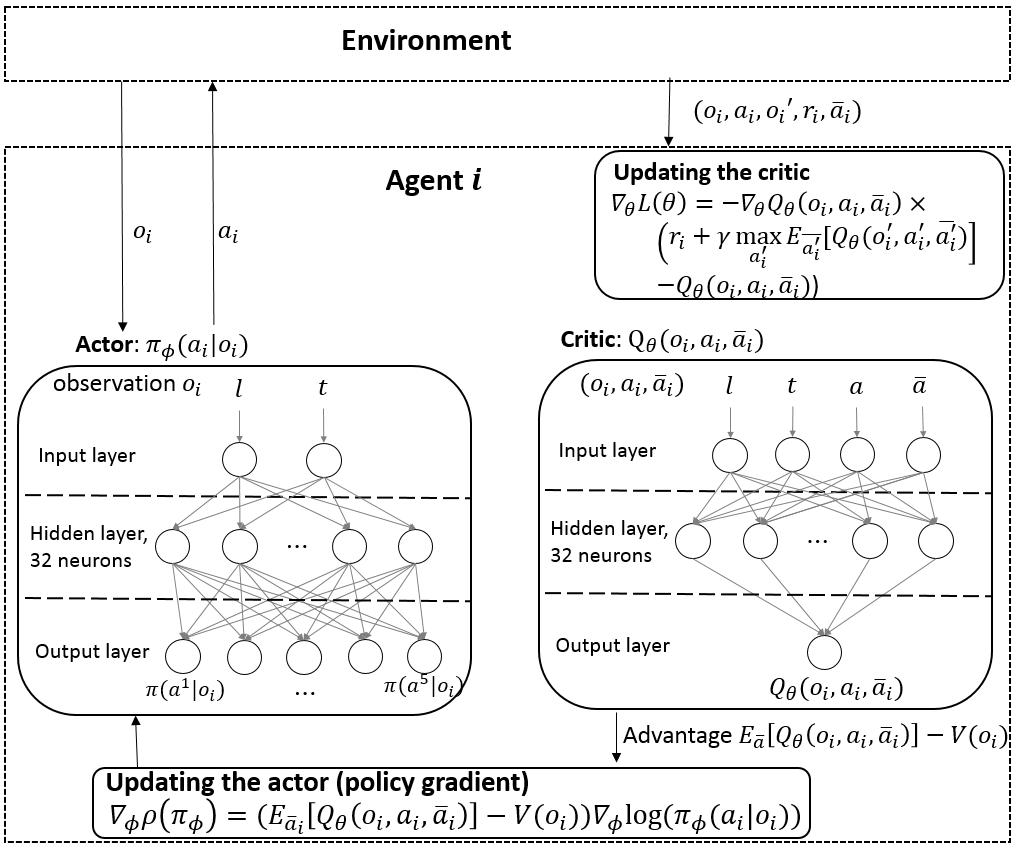}
	\centering 
	\caption{Mean field actor-critic algorithm for multi-driver repositioning}
	\label{fig:ac_mean}
\end{figure}

The derived Q values corresponding to four scenarios of interest are presented in Figure~(\ref{fig:cases}). In Figure~(\ref{subfig:case1}), when both drivers choose action \#4, the observed mean action for both of them is the ratio of demand to supply, i.e., $\bar{a}_1 = \bar{a}_2 = \dfrac{1~\text{passenger request}}{2~\text{drivers}} = 0.5$. The resulting expected value for both drivers is $\$3.5$, i.e., $Q(o_1, a_1, \bar{a}_1) = Q(o_2, a_2, \bar{a}_2) = 3.5$, because both of them have an equal probability $\dfrac{1~\text{driver}}{2~\text{drivers}} = 50\%$ to take the passenger request with $\$7$. Similarly, the observed mean actions and resulting Q values can be explained in other scenarios. 
The Q-value bimatrix is presented in Table~(\ref{tab:qvals}) where driver $1$ is the column player and driver $2$ is the row player. When driver $1$ chooses action $\#1$ and driver 2 chooses action $\#1$, Q-values for them are $1.5$ and $1.5$, respectively, according to Figure~(\ref{subfig:case4}). Similarly, Q-values for both drivers can be read from Figure~(\ref{fig:cases}) for other scenarios. Based on the bimatrix, driver $1$ always chooses action $\#4$ because action $\#4$ is strictly better than action $\#1$ regardless of the observed mean action, and driver $2$ always chooses action $\#4$ for the same reason. 
Thus, the optimal policy for both drivers is to enter grid $\#4$ with an expected payoff $\$3.5$. 
 
\begin{figure}[H]
	\centering 
	\subfloat[Both drivers take action \#4]{\includegraphics[scale=.85]{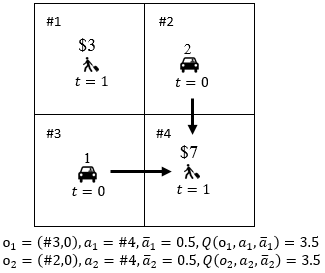}\label{subfig:case1}} ~
	\subfloat[Driver 1 takes action \#4 and driver 2 takes action \#1 ]{\includegraphics[scale=.85]{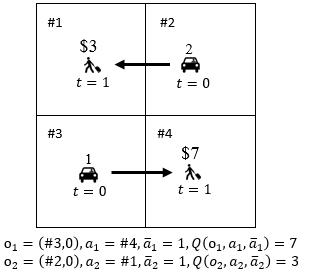}\label{subfig:case2}}
	\\
	\subfloat[Driver 1 takes action \#1 and driver 2 takes action \#4]{\includegraphics[scale=.85]{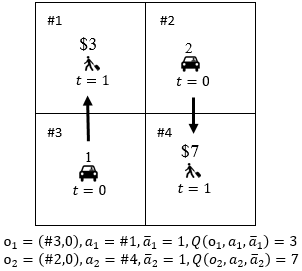}\label{subfig:case3}} ~
	\subfloat[Both drivers take action \#1 ]{\includegraphics[scale=.85]{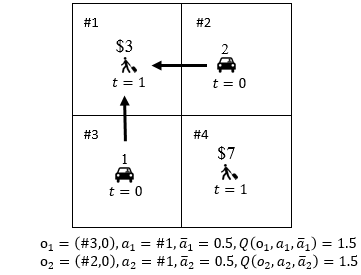}\label{subfig:case4}}
	\caption{Derived Q values for four scenarios of interest}
	\label{fig:cases} 
\end{figure}

\begin{table}[H]
	\centering\caption{Q-value bimatrix for drivers}
	\label{tab:qvals}
	\begin{tabular}{| p{35pt} p{7.5pt} p{7.5pt}| p{100pt} |p{100pt} |} 
		\hline
		 \multicolumn{3}{|c}{\multirow{2}{*}{}} & \multicolumn{2}{|c|}{Driver 1} \\ \cline{4-5}
		 & & & $\#1$ & $\#4$ \\ \cline{1-5}
		 \multirow{2}{*}{Driver 2} & \multicolumn{2}{|l|}{$\#1$} & $1.5, 1.5$ (Figure~\ref{subfig:case4})& $7,3$ (Figure~\ref{subfig:case2})\\ \cline{2-5}
		 & \multicolumn{2}{|l|}{$\#4$} & $3,7$ (Figure~\ref{subfig:case3})& $3.5,3.5$ (Figure~\ref{subfig:case1})\\ \hline
	\end{tabular}
\end{table}

\end{exmp}

\section{Reward design for multi-agent reinforcement learning}\label{sec:rd}








Due to selfishness of each agent, 
performing MARL under a given reward function in an MAS
is very likely to yield an undesirable equilibrium from the perspective of the system. 
In other words, this equilibrium may not be an optimum with respect to some system objectives.  
To guide an MAS towards a desirable equilibrium, system planners could resort to reward design mechanisms by modifying the reward function of agents. In this paper, we introduce a new parameter $\alpha \in \mathcal{A}$ into agents' reward, where $\mathcal{A}$ is the feasible domain of $\alpha$. Parameter $\alpha$ can be either a scalar or a vector. The goal of system planners is to maximize some system performance measure dependent of $\alpha$, denoted as $f(\alpha)$. 
The system planner first chooses a value of $\alpha$ and inputs it to the MAS. With the given $\alpha$ that influences the reward design, the developed mean field actor-critic algorithm is employed to derive an optimal policy $\pi$, which is dependent on $\alpha$, for all agents in the system. Some performance measure $f$, which is calculated by executing the derived optimal policy $\pi$ for all agents, is then fed into the reward design. The performance measure $f$ is dependent on $\alpha$ through the dependency of $\pi$ on $\alpha$. In other words, $f = f(\alpha)$.

In summary, the reward design problem is to select a parameter $\alpha$ to maximize the performance measure $f(\alpha)$ on the upper level, while the distributed agents aim to maximize their individual cumulative rewards on the lower level once $\alpha$ is given as part of their reward. 
This process can be formulated as a bi-level optimization problem, mathematically, 
\begin{align}
&max_{\alpha \in \mathcal{A}}f(\alpha) \label{eqn:max} \\
&~\text{where} \nonumber\\
& \qquad max_{\pi} \sum_{k = t}^T \gamma^{k - t}r_i^k(\alpha),~\forall t \in \{0,1,\cdots,T\},~\forall i \in \{1,2,\cdots,N\}. \label{eqn:max2} \nonumber
\end{align}

\begin{figure}[H]
	\centering
	\includegraphics[width=0.3\linewidth,height=0.2\textheight,keepaspectratio]{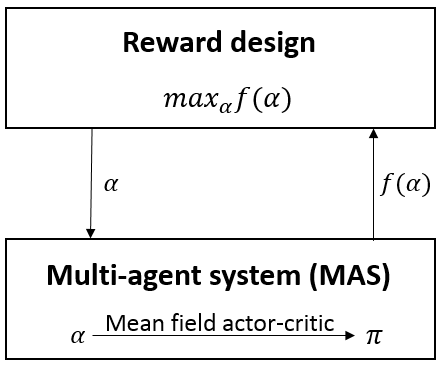}
	\centering 
	\caption{Architecture of the reward design}
	\label{fig:rd}
\end{figure}

The interaction between upper and lower levels through exchange of variables is shown in Figure~(\ref{fig:rd}).

The optimization problem presented in Equation~(\ref{eqn:max}), however, is not straightforward to solve due to the unknown complex structure of $f$ over the parameter $\alpha$. 
Traditional gradient based methods such as gradient descent are thus no longer applicable. In addition, the black-box function $f$ is expensive to evaluate, meaning that evaluating one value of $\alpha$ can take a long period of time. Therefore random or grid search based optimization techniques \citep{bergstra_random_2012} are less effective because they typically require a large number of evaluations. 

	To find optima with a limited number of evaluations, we adopt Bayesian optimization \citep{mockus_bayesian_1989} (hereafter we call it BO). Due to the unknown structure of the objective function $f$, BO places a prior on the objective function, for example a Gaussian process (GP).  With some observed data (i.e., several evaluations of $\alpha$'s), BO derives a posterior, based on which the next $\alpha$ to be evaluated can then be determined.


	We now detail the procedure of BO in our context. First, BO places a prior statistical model on the objective function $f$, for example a GP in this study. A GP is a stochastic process satisfing the following conditions: 1) For any $\alpha \in \mathcal{A}$, $f(\alpha)$ is a random variable; 2) For a collection of any finite number of $\alpha$'s (i.e., $\alpha_1, \alpha_2, \cdots,$ and $\alpha_m$ $\forall m \in \mathcal{R}$), the joint distribution $\boldsymbol{f} = [f(\alpha_1), f(\alpha_2), \cdots, f(\alpha_m)]^T$ is a multivariate Gaussian. Mathematically, 
\begin{equation*}
	\boldsymbol{f} \sim \mathcal{N}(\boldsymbol{\mu}, \boldsymbol{K}),
\end{equation*}
where $\boldsymbol{\mu}$ is a $m$ by $1$ mean vector and $\boldsymbol{K}$ is a $m$ by $m$ covariance matrix with each entry $K_{ij}$ denoting the covariance between $f(\alpha_i)$ and $f(\alpha_j)$. 
A square exponential kernel $\kappa(\alpha_i, \alpha_j)$ is used to calculate the covariance $K_{ij}$ and is defined as
\begin{equation*}
\kappa(\alpha_i, \alpha_j) = \sigma_f^2 e^{-\frac{1}{2l^2} (\alpha_i - \alpha_j)^2},
\end{equation*}
where $l$ is the length parameter and $\sigma_f$ the variance parameter. The square exponential kernel yields a large covariance when $\alpha_i$ and $\alpha_j$ are close (i.e., $(\alpha_i - \alpha_j)^2$ is small) and a small covariance when $\alpha_i$ and $\alpha_j$ are far away (i.e., $(\alpha_i - \alpha_j)^2$ is large), which is as expected because for a typical function $f$, $f(\alpha_i)$ and $f(\alpha_j)$ are similar when $\alpha_i$ and $\alpha_j$ are similar, meaning a large covariance. The prior used in this study is a GP with a zero mean (i.e., $\boldsymbol{\mu} = \boldsymbol{0}$) and a covariance matrix $\boldsymbol{K}$ calculated by the aforementioned square exponential kernel.

Second, with a given prior and some observed data, now our task is to derive a posterior probability distribution of $\boldsymbol{f}^* = [f_1^*, f_2^*, \cdots, f_m^*]^T$ at some $\boldsymbol{\alpha}^* = [\alpha_1^*, \alpha_2^*, \cdots, \alpha_m^*]^T$. We assume $\boldsymbol{f} =[f_1, f_2, \cdots, f_n]^T$ is the observed data at some $\boldsymbol{\alpha} = [\alpha_1, \alpha_2, \cdots, \alpha_n]^T$. By the definition of a GP, the joint distribution $(\boldsymbol{f}, \boldsymbol{f}^*)$ is a multivariate Gaussian. Mathematically, the prior probability distribution of  $(\boldsymbol{f}, \boldsymbol{f}^*)$ with zero mean is 
\begin{equation*}
\begin{bmatrix}
\boldsymbol{f}\\
\boldsymbol{f}^*
\end{bmatrix} \sim \mathcal{N}
\left (
\begin{bmatrix}
\boldsymbol{0}_n \\
\boldsymbol{0}_m
\end{bmatrix}, 
\begin{bmatrix}
 \boldsymbol{K}_{n\times n} & \boldsymbol{K}^*_{n\times m}\\
\boldsymbol{K}^{*T}_{m\times n} & \boldsymbol{K}^{**}_{m \times m}
\end{bmatrix}
\right ), 
\end{equation*}
where $\boldsymbol{0}_n$ is $n$-dimensional vector with all entries as zeros, $\boldsymbol{0}_m$ $m$-dimensional vector with all entries as zeros, $\boldsymbol{K}$ a $n$ by $n$ covariance matrix with each entry $K_{ij} = \kappa(\alpha_i, \alpha_j) + \sigma_y^2 \delta_{ij}$ ($\sigma_y$ is the noise in the observed data and $\delta_ij$ is the Kronecker delta), $\boldsymbol{K}^*$ a $n$ by $m$ matrix with each entry $K^*_{ij} = \kappa(\alpha_i, \alpha_j^*)$, and $\boldsymbol{K}^{**}$ a $m$ by $m$ matrix with each entry $K^{**}_{ij} = \kappa(\alpha_i^*, \alpha_j^*)$. According to the multivariate Gaussian theorem \citep{murphy_machine_2012}, the probability distribution of $\boldsymbol{f}^*$ conditioned on the observed $\boldsymbol{f}$ is 
\begin{equation}
p(\boldsymbol{f}^*| \boldsymbol{f}) \sim \mathcal{N}\left( \boldsymbol{\mu}^*, \boldsymbol{\Sigma}^* \right),
\label{eqn:post}
\end{equation}
where $\boldsymbol{\mu}^* = \boldsymbol{K}^{*T}\boldsymbol{K}^{-1}\boldsymbol{f}$ and $\boldsymbol{\Sigma}^* = \boldsymbol{K}^{**}- \boldsymbol{K}^{*T} \boldsymbol{K}^{-1} \boldsymbol{K}^*$.

	Third, with the derived posterior probability distribution of $\boldsymbol{f}^*$ conditioned on the seen data $\boldsymbol{f}$, we now aim to define an acquisition function, based on which the next $\alpha$ to be evaluated can be determined. The acquisition function used in this study is the upper confidence bound (UCB) \citep{srinivas_gaussian_2010}, i.e., 
\begin{equation}
	UCB(\alpha) = \mu^*(\alpha) + \sqrt{2log(t^{d/2+2}\pi^2/3\delta)} \sigma^*(\alpha),
	\label{eqn:ucb}
\end{equation}
where $\mu^*(\alpha)$ is the posterior mean at $\alpha$, $\sigma^*(\alpha)$ the posterior standard deviation at $\alpha$ (derived from $\boldsymbol{\Sigma}^*$), $d$ is the dimension of the search space of $\alpha$, and $\delta$ is a parameter. 
The next $\alpha'$ to be evaluated is simply 
\begin{equation}
	\alpha' = \argmax_{\alpha \in \mathcal{A}} UCB(\alpha).
	\label{eqn:next}
\end{equation}
\begin{rem}
	With the acquisition function defined as $UCB$ in Equation~(\ref{eqn:ucb}), it has been shown that BO is no regret with high probability and has a lower-bound on the convergence rate \citep{brochu_tutorial_2010}. Interested readers are referred to \cite{srinivas_gaussian_2010, brochu_tutorial_2010, berkenkamp_no-regret_2019} for more details on global optimality and convergence properties of BO. 
\end{rem}	
The pseudo-code of BO used in this study is listed in Algorithm~(\ref{alg:bayesian}).


\begin{algorithm}[H]
	\caption{Bayesian Optimization}
	\label{alg:bayesian}
	\begin{algorithmic}[1]
		\State Initialize a GP prior with zero mean on the objective function $f$
		\State Set computational budget $\mathcal{B}$, a initial number of evaluations $b_0$, and $b = b_0$
		\State Evaluate $f$ at $b_0$ randomly chosen $\alpha$'s, i.e., $\boldsymbol{f} = [f(\alpha_1), f(\alpha_2), \cdots, f(\alpha_{b_0})]$
		\While {$b \leq \mathcal{B}$}
		\State Calculate the posterior probability distribution of $\boldsymbol{f}^*$ conditioned on $\boldsymbol{f}$ by Equation~(\ref{eqn:post})
		\State Calculate the acquisition function $UCB$ by Equation~(\ref{eqn:ucb}) with $t = b$ 
		\State Locate the next $\alpha'$ to be evaluated by Equation~(\ref{eqn:next})
		\State  Evaluate $f$ at $\alpha'$ and append $f(\alpha')$ to $\boldsymbol{f}$
		\State $b \leftarrow b + 1$
		\EndWhile
		\State Return $\alpha$ that maximizes $f$
	\end{algorithmic}
\end{algorithm}


To be more concrete, now we use the multi-agent $2 \times 2$ example presented in Figure~(\ref{subfig:toy_example}) to illustrate the potential of the reward design. 
\begin{exmp} (Multi-Agent $2 \times 2$).  
	We take the order response rate (ORR), i.e. the ratio of the number of fulfilled passenger requests to the total number of passenger requests, as the performance measure of the system.  
	The direct application of mean field actor-critic algorithm 
	yields a 50\% ORR, which is obviously not the desired equilibrium from the perspective of the system. Noticing that the platform typically charges a certain proportion of the fare paid by the passenger as the so-called platform service charge, which is reportedly to be dependent on various factors such as distance, duration, and city. We aim to improve the performance of the system by devising a proper reward design. 
	
	In Figure~(\ref{subfig:toy_example}), trip fares are shown right above each passenger request, the reached equilibrium for both drivers without any charge are to enter grid $\#4$ and get an expected reward as $\$3.5$, leading to an oversupply (i.e., a low demand to supply ratio) in grid $\#4$ and an undersupply (i.e., a high demand to supply ratio) in grid $\#1$, which is not beneficial for the system. A reward design which deducts $\$1.1$ from the passenger request paid to the driver in grid $\#4$ will effectively attract one driver to leave grid $\#4$ for grid $\#1$ to get more monetary return, resulting in a $100\%$ order response rate. 
	\label{exmp1}
\end{exmp}

\section{Case Study}\label{sec:case}

To test the performance of the proposed bilevel optimization model, we use two datasets including a synthetic dataset and one real-world large-scale taxi dataset downloadable from official website of New York City (NYC) Taxi \& Limousine Commission (https://www1.nyc.gov/site/tlc/about/tlc-trip-record-data.page).  

\subsection{E-hailing driver repositioning under service charge}
We first test the bilevel optimization model on a 2-by-2 grid world example, where an analytical solution of the reward design can be derived. Then we compare both values to justify the correctness of our BO algorithm. 

\subsubsection{Lower level MARL setup}
\begin{figure}[H]
	\centering
	\includegraphics[width=0.5\linewidth,height=0.25\textheight,keepaspectratio]{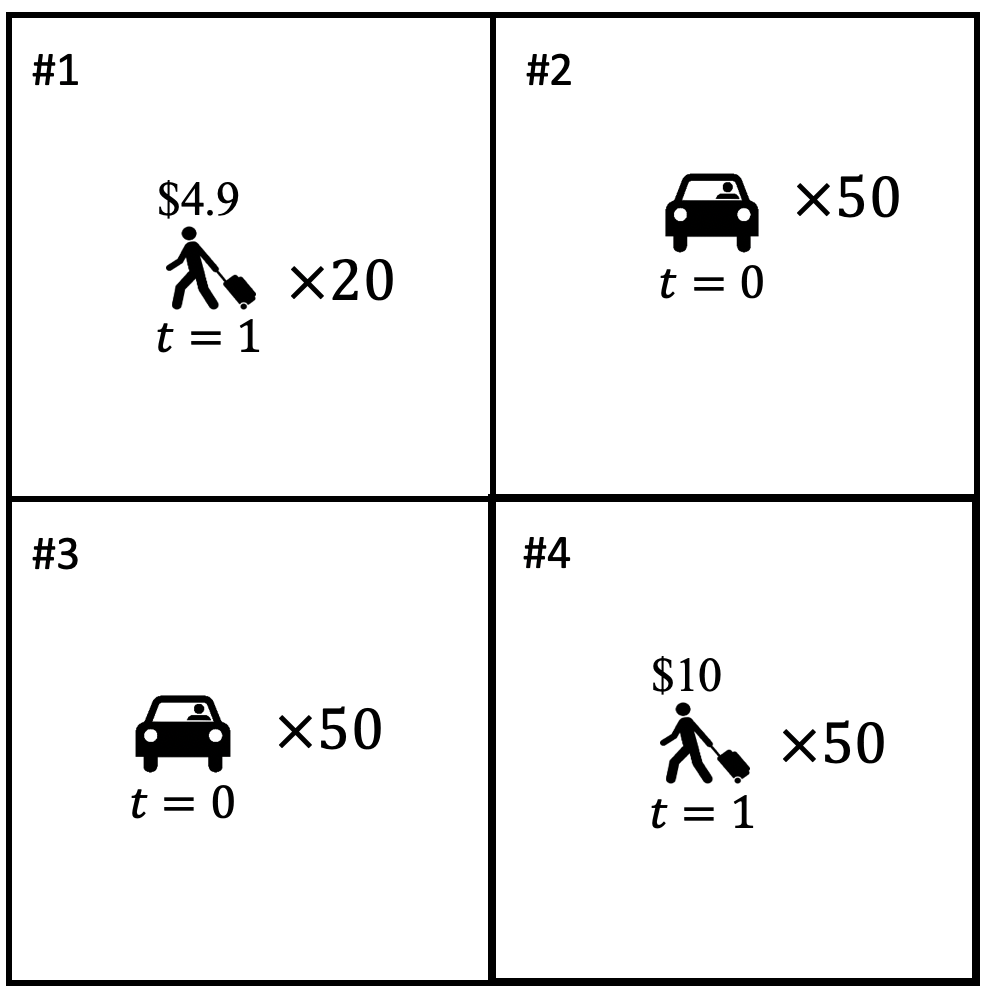}
	\centering 
	\caption{Layout}
	\label{fig:small_setup}
\end{figure}

The dataset consists of seven deterministic passenger requests in a 2-by-2 grid world setup, as shown in Figure~(\ref{fig:small_setup}). At $t = 0$, there are fifty idle drivers in grid $\#2$ and fifty in grid $\#3$. At time $t=1$, fifty passenger requests with fare $\$10$ deterministically appear in grid $\#4$ and twenty passenger requests with fare $\$4.9$ appear in grid $\#1$. The observation space for driver $i$ consists of the grid index and current time, i.e., $o_i = (l_i, t)$, and the action space is to enter one of neighboring grids or to stay at the current grid. The reward that a driver earns is her monetary return. That is, $$r = \text{fare}\times (1 - SC_l),$$ where $SC_l$ is the service charge in percentage if the driver takes the passenger request in grid $l$. $SC_l$ is defined later in Equation~(\ref{eqn:sc}). 

\subsubsection{Upper level objective function}

Without any reward design, the optimal policy for all drivers is to enter grid $\#4$, because the expected return for entering grid $\#4$ is at least $\$5$ (i.e., 100 drivers compete for 50 orders with $\$10$ each) while that for entering grid $\#1$ is at most $\$4.9$ (i.e., the highest fare of an order in $\#1$ is $\$4.9$).  The resulting ORR is $50/70 = 71.43\%$, which is not desirable from the perspective of the platform because it is expected to achieve a $100\%$ ORR in this setting. Actually, the platform can achieve a better ORR by adjusting the reward that drivers earn through the use of a platform service charge (aka the commission fee). The platform service charge used in this study is denoted as a fare percentage. For instance, a 10\% service charge means the platform takes 10\% of the fare paid by the passenger to the driver as its revenue. In other words, the driver gets less money under a higher service charge while the payment from the passenger remains the same. To achieve a better ORR, the platform needs to place a high service charge in grids which are oversupplied. Drivers oversupply grid $l$ because on average they can earn more by entering grid $l$, compared with entering other grids. A high service charge placed in grid $l$ can effectively reduce monetary returns for drivers entering $l$ and make grid $l$ less attractive to drivers. Thus, some drivers choose other grids and take other passenger requests, resulting in an increase in ORR.  

Before introducing a functional form of the platform service charge, we formly provide two notations, namely demand to supply ratio (DS) and service charge (SC). We then construct an effective form of SC as a function of DS. A small DS indicates that the grid is oversupplied, and a large DS means the grid is undersupplied. The goal of the platform is to drive DS close to $1$, meaning a balance between demand and supply. In a grid $l$ with $DS_l$ below $1$, $SC_l$ is expected to be large to discourage drivers from oversupplying the grid; while in a grid $l$ with $DS_l$ above $1$, $SC_l$ is supposed to be small. To illustrate such a relation, we use a piecewise linear function with a parameter $\alpha$ as SC in grid $l$, i.e.,
\begin{equation}
SC_l =
\begin{cases}
\alpha \times (1 - DS_l) & \text{if}~DS_l \leq 1,\\
0 & \text{otherwise,}
\end{cases}       
\label{eqn:sc}
\end{equation}
where a relatively high SC is charged to all drivers in the grid with a low DS and no SC is charged to drivers in the grid with DS above $1$. Parameter $\alpha$ is the SC at $DS_l = 0$. 

With an adjustable parameter $\alpha$, the platform aims to maximize some objective $f$ consisting of two components, namely ORR and overall service charge (OSC), as 
\begin{equation}
f = w \times \text{ORR} + (1-w) \times (1 - \text{OSC}).
\label{eqn:small_objective}
\end{equation}
OSC is defined as
\begin{equation*}
OSC = \dfrac{\sum_l \sum_{\text{order} \in \text{serviced\_orders}}  \mathbbm{1}_{\text{order}_{\text{origin}} \in l} \times SC_l \times \text{fare}_{\text{order}}}{\sum_{\text{order} \in \text{serviced orders}} \text{fare}_{\text{order}}},
\end{equation*}
where the denominator is the total amount of fare of all serviced orders and the numerator is the total amount of service charge. 

The rationale of choosing these two components is as follows. First, from the perspective of the platform, it aims to maximize ORR, because a larger ORR typically means a higher revenue and a higher customer satisfaction. To maximize ORR, the platform simply chooses the largest possible value of $\alpha$. 
The reason is that with the largest possible $\alpha$, the platform penalizes drivers heavily for oversupplying a grid, and therefore drivers will be directed to other grids. This strategy, i.e., choosing the largest $\alpha$, however, is a big threat for the long-term growth of the platform because drivers are very likely to quit under such a high service charge. Thus, the platform also needs to maintain a relatively small OSC. 
Considering the competition between ORR and OSC, we use a weighted average of ORR and $(1-\text{OSC})$ as the objective of the platform, i.e., 
where $w \in [0,1]$ is the weight for $\text{ORR}$. In this case study, we set $w = \frac{3}{5}$, meaning that the platform cares more about ORR.

\subsubsection{Results}
We then use two methods, namely BO and an analytical method, to determine the optimal value of $\alpha$. 

\begin{enumerate}
	
    \item \textbf{BO}. 
	We first employ BO with the objective function given in Equation~(\ref{eqn:small_objective}). 
	When we apply the mean field actor-critic algorithm to the lower level MAS, the critic is parameterized by a multilayer perceptron (MLP) with three hidden layers $(64,32,16)$, and the actor is parameterized by an MLP with three hidden layers $(32,16,8)$.  ReLU is used as the activation function between hidden layers, and a softmax function is used to transform the final output from the actor to be a probability distribution. Learning rates $\eta$ for both the actor and the critic are $10^{-4}$. Discount factor $\gamma = 1$ under the assumption that drivers typically do not depreciate their monetary income on the same day. Exploration parameter is initially set as $\epsilon_0 = 0.5$ and linearly decreases over iterations until it reaches a minimum value of $0.01$. The target network update period is $\tau = 10$, meaning that the parameters of target networks are updated every $10$ episodes.  
	
	For a bilevel optimization problem, first we need to check the convergence of the lower level. As an example to validate the convergence, ORR, (1 - OSC), and the average reward of all drivers (i.e., Reward in Figure~(\ref{fig:convergence})) versus the index of episodes are presented in Figure~(\ref{fig:convergence}) with $\alpha = 0.5$ and $\alpha=1.0$. In both scenarios, ORR and reward increase very fast and (1 - OSC) steadily decreases during the first 500 episodes where agents explore the environment and learn the optimal policy. ORR, (1 - OSC), and reward gradually converge after 1,500 episodes when agents mainly exploit the knowledge they have gained through their previous explorations.

	\begin{figure}[H]
		\centering 
		\subfloat[$\alpha=0.5$]{\includegraphics[scale=.45]{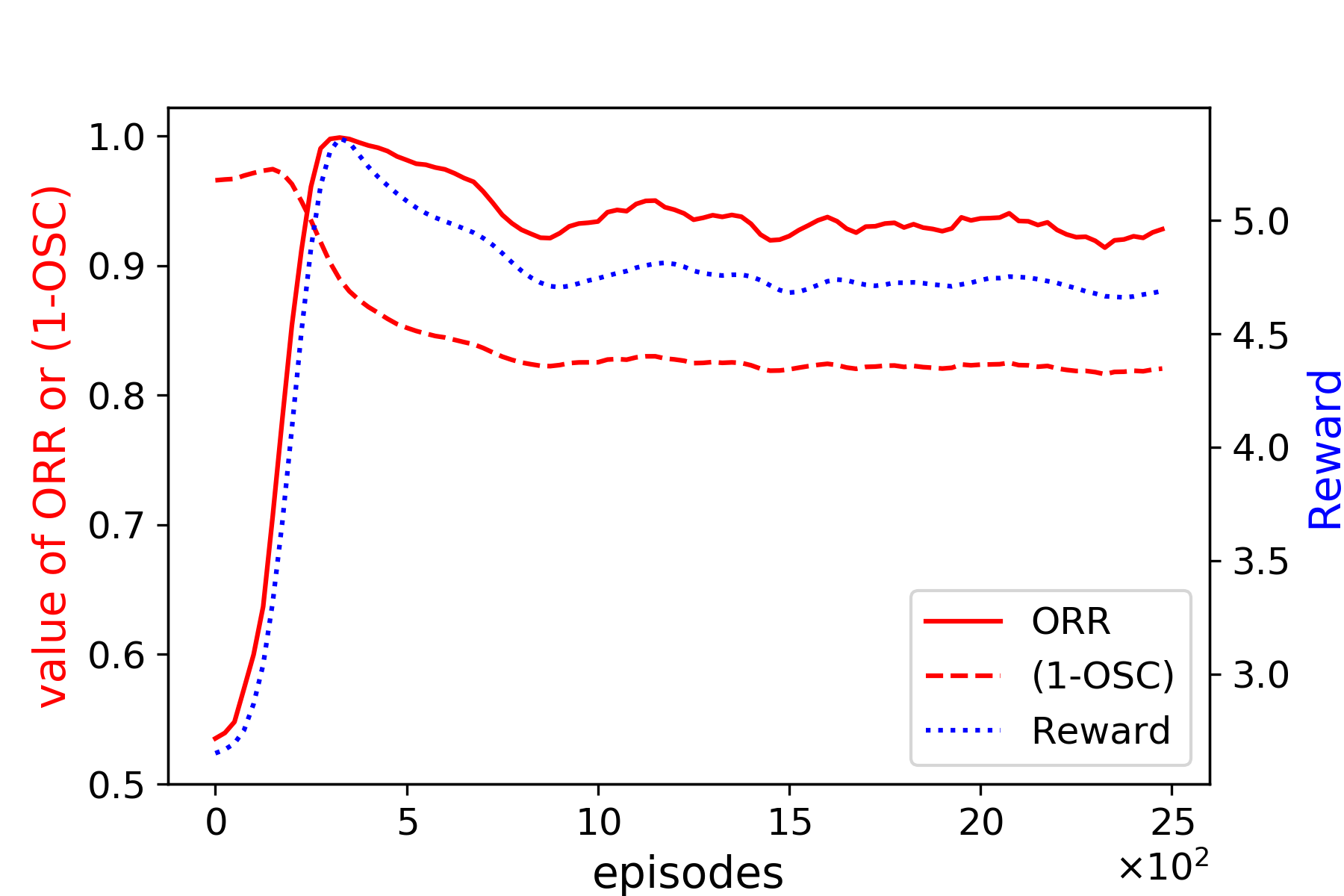}\label{subfig:alpha05_convergence}} ~
		\subfloat[$\alpha=1.0$]{\includegraphics[scale=.45]{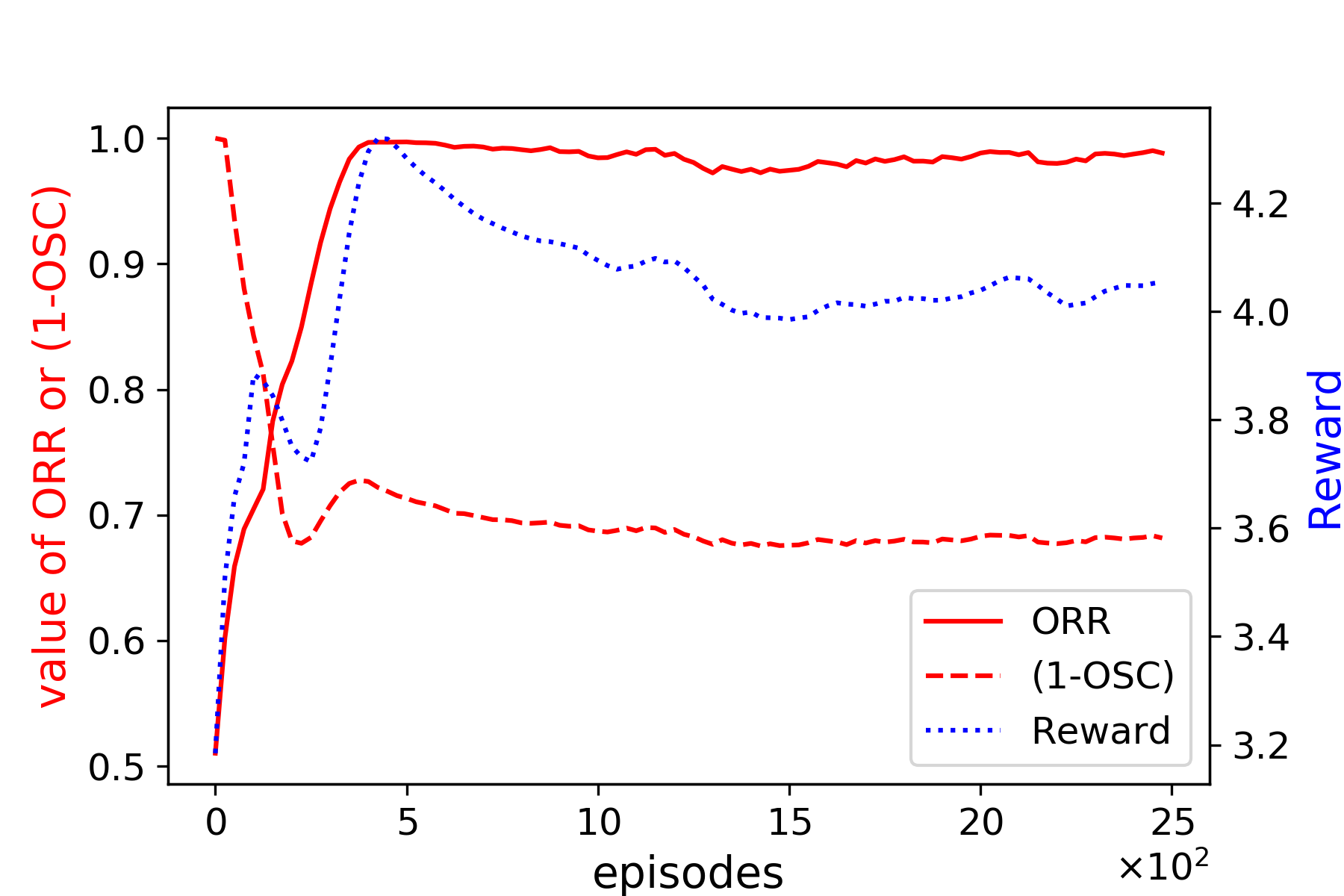}\label{subfig:alpha1_convergence}}
		\caption{Convergence of the lower level MARL}
		\label{fig:convergence} 
	\end{figure}


	\begin{figure}[H]
		\centering 
		\subfloat[Posterior probability distribution of $f$ conditioned on the initial five evaluated points]{\includegraphics[scale=.5]{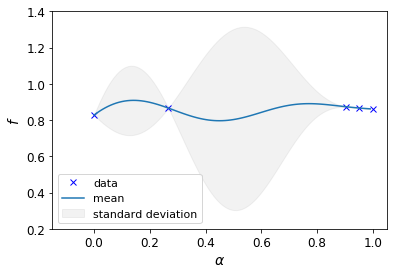}\label{subfig:BO_start}} ~
		\subfloat[Acquisition function ($UCB$)]{\includegraphics[scale=.5]{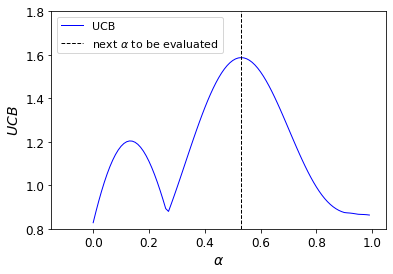}\label{subfig:acqui}}
		\caption{Posterior probability distribution and acquisition function at iteration $0$}
		\label{fig:BO} 
	\end{figure}

	With the validated convergence of the lower level MARL, we now run BO to solve the bilevel optimization problem. 
	As for the initial value of $\alpha$, in addition to $0$ and $1$ (i.e., both ends of the interval of interest), we randomly sample three other values according to a uniform distribution. In total we evaluate five $\alpha$'s as the starting point of BO. Figure~(\ref{subfig:BO_start}) plots the mean and standard deviation of the posterior probability distribution of $f$ conditioned on these five evaluated points. As one can see, the standard deviation is small around the points that have been evaluated and is large at locations where we do not have any data. The acquisition function shown in Figure~(\ref{subfig:acqui}) reveals that the next $\alpha$ to be evaluated is around $0.53$. 

	
	\begin{figure}[H]
		\centering
		\includegraphics[width=0.99\linewidth,height=0.2\textheight,keepaspectratio]{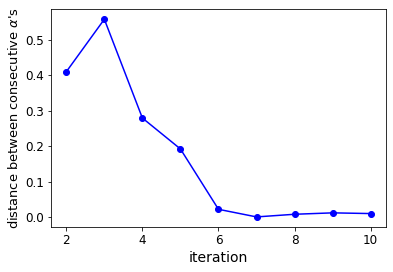}
		\centering 
		\caption{Convergence of BO}
		\label{fig:convergence_bo_small}
	\end{figure}


	With the initial five evaluated points, we run BO until convergence. The convergence of BO in this study is defined as choosing 5 consecutive $\alpha$'s with the distance between every two consecutive $\alpha$'s below a threshold of $0.05$. In other words, BO converges when it starts choosing similar $\alpha$'s to evaluate. The convergence is presented in Figure~(\ref{fig:convergence_bo_small}).  Note that the horizontal axis starts with iteration index $2$ because it is meaningless to calculate the distance between $\alpha$ in the first iteration and the initial five $\alpha$'s. We can see that BO initially chooses quite different $\alpha$'s and gradually converges after the $6^{th}$ iteration. In other words, BO chooses similar $\alpha$'s after the $6^{th}$ iteration. 
	
	The resulting posterior probability distribution of $f$ from BO is presented in Figure~(\ref{fig:BO_final}). It is noticeable that the evaluation of the objective on $\alpha$'s seems noisy. In other words, the evaluated objective may be slightly different at the same $\alpha$. This is expected because there are multiple local optima when solving the lower level MARL. Actually, it is commonly impossible to find a global optimum using deep learning. Thus researchers usually settle for local optima \citep{Goodfellow-et-al-2016}. Local optima introduce noise into the evaluation of the objective at each $\alpha$. Although the evaluations are noisy, the fitted curve is able to capture the mean objective $f$ for each $\alpha$. The optimal $\alpha$ is determined as $0.55$, yielding an optimal mean objective $f = 0.90$. The optimum is 8.4\% higher than the objective $f = 0.83$ without any reward design.  
	

	\begin{figure}[H]
		\centering
		\includegraphics[width=0.79\linewidth,height=0.25\textheight,keepaspectratio]{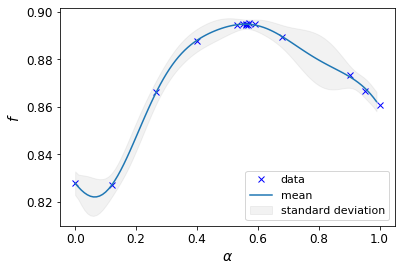}
		\centering 
		\caption{Posterior probability distribution of $f$ after BO converges}
		\label{fig:BO_final}
	\end{figure}
	
	\item \textbf{Analytical method}. 
	Due to the simplicity of this case, we can analytically derive the optimal value of $\alpha$ and shed some light on the effectiveness of the proposed platform service charge. Recall that the optimal policy for all drivers is to enter grid $\#4$ when $\alpha = 0$. The resulting DS ratio in grid $\#4$ is $5/10 = 0.5$, which is well below $1$, meaning that grid $\#4$ is oversupplied. ORR is $5/7 = 71.4\%$. To increase ORR, one needs to increase $\alpha$ to penalize drivers who oversupply a grid. As $\alpha$ gradually increases, grid $\#4$ becomes less attractive, because the expected return one driver can earn decreases as $\alpha$ increases. When the expected return one driver can earn is less than $\$4.9$, some drivers will choose to enter grid $\#1$ instead of grid $\#4$ for a higher monetary return. 
	Note that to ease the analysis, we assume the number of drivers entering a grid is always an integer. 
	Now we present how we calculate the critical value of $\alpha$ below which there is no driver choosing to enter grid $\#1$ while above which there is one driver attracted by grid $\#1$. With one driver entering grid $\#1$, there are 99 drivers entering grid $\#4$, resulting a $50/99 = 0.51$ DS ratio in grid $\#4$. $SC_{\#4} = \alpha \times (1 - 0.51) = 0.49 \alpha$, meaning that the expected return for these 99 drivers is $\dfrac{\$10 \times (1 - 0.49\alpha) \times 50}{99} = 5.05 \times (1 - 0.49 \alpha)$. The expected return for the driver entering grid $\#1$ is $\$4.9$. We then have the critical condition $5.05 \times (1 - 0.49 \alpha) = 4.9$, yielding $\alpha = 0.06$. Similarly, we calculate the critical value of $\alpha$ below which there are 19 drivers choosing to enter grid $\#1$ while above which there are 20 drivers attracted by grid $\#1$, and the critical value is $\alpha = 0.58$. 
	
	\begin{table}[H]
		\centering\caption{Values of interest}
		\label{tab:small_objective}
		\begin{tabular}{| p{130pt} | p{95pt} |p{65pt} | p{65pt}|} 
			\hline
			$\alpha$ & $0$ & $0.06$ & $0.58~(\text{optimal})$ \\ \hline
			DS ratio in grid $\#1$ ($t=1$) & N.A. (supply is zero) & $20/1=2000\%$ & $20/20=100\%$ \\ \hline
			DS ratio in grid $\#4$ ($t=1$) & $50/100 = 50.0\%$ &  $50/99 = 50.5\%$ & $50/80 = 62.5\%$\\ \hline
			ORR & $50/70 = 71.4\%$ &  $51/70 = 72.9\%$ & $100\%$\\ \hline
			OSC & $0$ & $0.03$ & $0.18$ \\ \hline
			$f$ & $0.83$ & $0.83$  & $0.93~(\text{optimum})$\\ \hline
		\end{tabular}
	\end{table}
	
	Values of interest are presented in Table~(\ref{tab:small_objective}). With $\alpha = 0$, $\text{ORR} = 71.4\%$ and $\text{OSC} = 0$. The objective is $\frac{3}{5} \times \text{ORR} + \frac{2}{5}(1 - \text{OSC}) = 0.83$. With $\alpha$ increasing to $0.06$, there is one driver attracted by grid $\#1$, resulting in a $72.9\%$ ORR. The OSC is calculated as follows. The DS ratio in grid $\#4$ is now $0.51$, resulting in $SC_{\#4} = 0.06 \times (1 - 0.51) = 0.03$. Thus, $\text{OSC} = \dfrac{\$10 \times 50 \times 0.03}{\$10 \times 50 + \$4.9} = 0.03$. The objective is $\frac{3}{5} \times \text{ORR} + \frac{2}{5}(1 - \text{OSC}) = 0.83$. Similarly, with $\alpha$ increasing to $0.58$, $\text{ORR} = 100\%$, $\text{OSC} = 0.18$, and the objective is $0.93$. Increasing $\alpha$ further does not improve ORR but increases OSC, resulting in a decrease in the objective. Thus, the analytically derived optimal value of $\alpha$ is $0.58$. 

\end{enumerate}

The derived optimal value of $\alpha$ from BO, i.e., $0.55$, agrees well with the analytically derived optimal value of $\alpha$, i.e., $0.58$. The optimum from BO, i.e., $0.90$ is quite close to its analytical counterpart, i.e., $0.93$. The small discrepancy between the numerical solution and the analytical one is explained as follows. In the analytical solution, the policy for agents is deterministic and exactly twenty drivers choose grid $\#1$ after increasing $\alpha$ to $0.58$; while in BO, the derived optimal policy for agents is stochastic, introducing variance in drivers' actions. For example, the derived optimal policy says each driver has a $20\%$ probability of choosing grid $\#1$ and a $80\%$ probability of choosing grid $\#4$. 
Although the expected number of agents in grid $\#1$ is $20$ and the expected number of agents in grid $\#4$ is $80$,  
the probability of 21 agents choosing grid $\#1$ is ${100 \choose 21}\cdot0.8^{79}\cdot 0.2^{21}=9.5\%$. This variance reduces both ORR and (1 - OSC), resulting in a lower objective from BO, compared with the objective from the analytical solution. 

\subsection{Multiclass taxi driver repositioning under congestion pricing}

In this case study, we apply the proposed bilevel optimization model to a real-world scenario where city planners aim to mitigate traffic congestion in the central business district (CBD). As an effective way to improve traffic condition, congestion pricing has been adopted by many cities such as London and Stockholm \citep{de_palma_traffic_2011}. The basic idea of congestion pricing is to impose a toll charge on all vehicles entering the CBD.  Consequently, some drivers may be sensitive to the toll charge and take alternative travel modes such as subway while some drivers could bear the toll charge. To demonstrate the effectiveness of congestion pricing, we use NYC taxi and subway data due to data availability. 

In the taxi market, 
congestion pricing affects both the demand and supply. 
On the demand side, the toll charge is passed to passengers for taxi drivers carrying passengers into the CBD. In other words, the fare paid by passengers is increased and thus the demand (i.e., number of passenger requests) is decreased. According to \cite{schaller_elasticities_1999}, taxi demand falls by $0.22 \times x$ percent when taxi fare increases by $x$ percent. For example, when taxi fare increases from $\$10$ to $\$12.5$ (i.e., increases by $25\%$) for a trip, 
the demand falls by $5.5\%$ (i.e. $0.22 \times 25\%$). On the supply side, the toll charge is paid by taxi drivers when they enter the CBD vacantly, which discourages drivers from entering the CBD without any passenger. 
The overall effect of congection pricing results in a reduced number of taxis in the CBD, leading to an improved traffic condition. This, however, may direct too many passengers, whose taxi requests are unfulfilled, to the public transit which is already running at full pressure during rush hours \citep{amy_2020}. Thus, there exists a tradeoff between reducing the number of taxis in the CBD and maintaining a reasonable level of crowdedness in the public transit system. 



\emph{Note.} Considering that the goal of this case study is to demonstrate the effectiveness of the proposed bi-level optimization model on a real-world large-scale problem, we simplify the modeling of demand to a linear model in which demand falls when fare increases. We acknowledge that a more realistic econometric model such as a logit model could be adopted to model passengers' willingness to pay and demand elasticity under congestion pricing. However, we may not be able to calibrate the econometric model due to data accessibility. In addition, the calibration process may introduce more uncertainty into our model. Therefore, a linear model is used as a demonstration in this case study, and a more realistic model to capture the coupling effect between the demand and supply is left for future research.

The objective of city planners thus consists of two components, namely, the number of taxis in the CBD and the crowdedness of the public transit system. Considering the accessibility of data (i.e., NYC taxi data and subway data), we make two assumptions: (1) The proposed congestion pricing scheme only affects the behavior of taxi drivers, as previously mentioned; 
(2) The subway system is used as a proxy of the public transit of the city. 

\begin{figure}[H]
	\centering
	\includegraphics[width=0.99\linewidth,height=0.15\textheight,keepaspectratio]{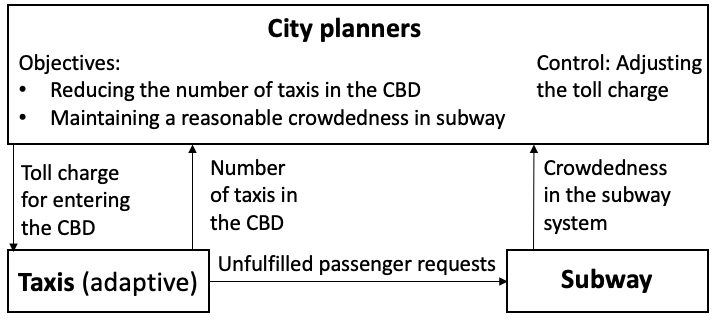}
	\centering 
	\caption{Objectives of city planners}
	\label{fig:cityplanners}
\end{figure}

 
To be more precise, Figure~(\ref{fig:cityplanners}) presents objectives of city planners and how planners derive the best control. City planners impose a toll charge on taxis entering the CBD. Adaptive taxis learn the optimal policy by the mean-field actor-critic algorithm under the toll charge. With fewer taxis searching for passengers in the CBD, more unfulfilled passenger requests are directed to the subway system. City planners observe the number of taxis in the CBD and crowdedness in the subway and adjust the toll charge to achieve a better balance between these two objectives. This process repeats until city planners reach a satisfactory balance. 
 
\subsubsection{Data preprocessing} \label{sec:preprocessing}

The NYC taxi trip records are publicly available on the official website of NYC Taxi \& Limousine Commission (https://www1.nyc.gov/site/tlc/about/tlc-trip-record-data.page). We use the data for both yellow and green taxis during May 2014 before the wide adoption of ridesharing service such as Uber and Lyft and after the business of green taxis gradually stabilizes. 
A data sample is listed in Table~(\ref{tab:data_sample}). Each entry in Table~(\ref{tab:data_sample}) collects the order information, including pickup and dropoff time and locations and fares (including tips). In total there are around 16 million taxi trips. We first remove the weekend data because trip patterns over weekends are obviously different from that on weekdays. We then restrict the time interval of interest as the evening peak, i.e., 4 PM to 8 PM. There are 2 million taxi trips in the weekday data after preprocessing. 
\begin{table}[H]
	\centering\caption{Taxi data sample}
	\label{tab:data_sample}
	\begin{tabular}{| p{55pt} | p{55pt} |p{50pt} | p{50pt}| p{50pt}| p{50pt} | p{35pt}|} 
		\hline
		pickup datetime & dropoff datetime & pickup longitude & pickup latitude & dropoff longitude & dropoff latitude & fare   \\ \hline
		2014-05-01 16:59:00 &  2014-05-01 17:08:30  & -73.978818 & 40.785048 & -73.965570 & 40.800718 & 6.5 \\ \hline 
		2014-05-01 16:59:00 & 2014-05-01 17:23:00 & -73.960280 & 40.778892 & -73.975542	 & 40.751427 & 15.5 \\ \hline
	\end{tabular}
\end{table}

\begin{figure}[H]
	\centering
	\includegraphics[width=0.99\linewidth,height=0.4\textheight,keepaspectratio]{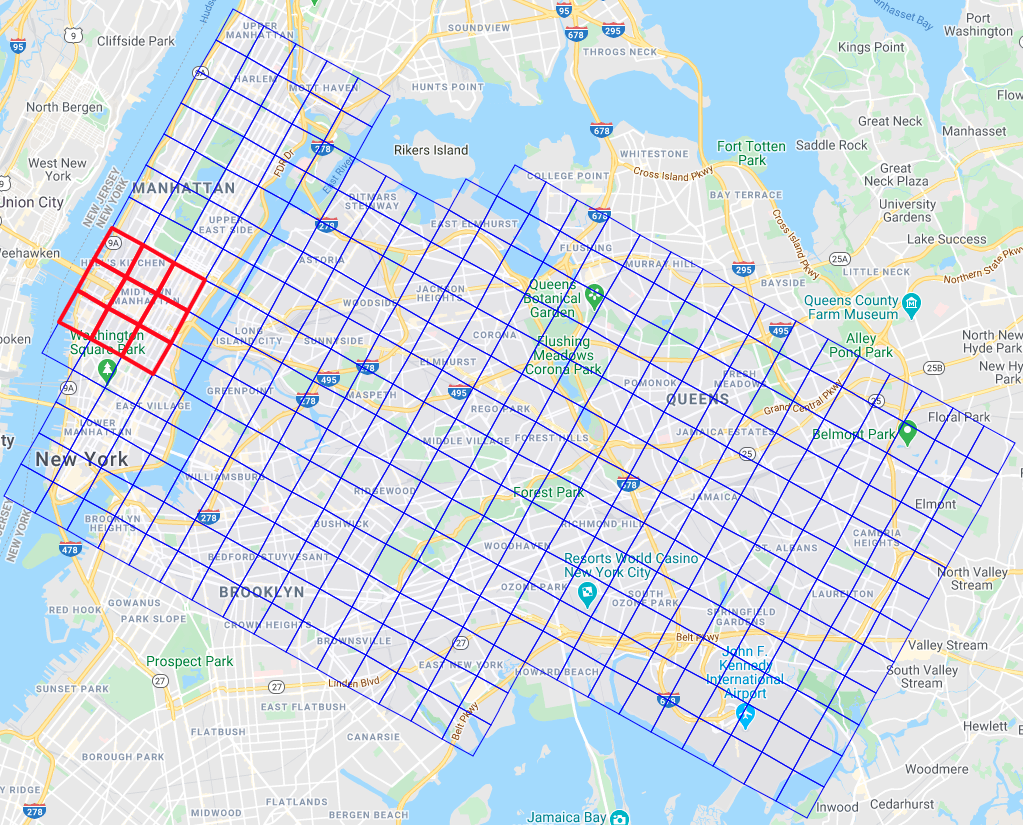}
	\centering 
	\caption{Spatial discretization of the area}
	\label{fig:study_area}
\end{figure}

Figure~(\ref{fig:study_area}) presents the spatial discretization of the area of interest. There are in total $337$ grids with a side length of 1 km covering the area from Manhattan to two airports located in Queens. Taxi orders outside grids consist of less than 2\% of the overall taxi orders and are not considered. Each longitude and latitude coordinate is transformed into a grid index. As for the temporal discretization, the evening peak is divided into eighty 3-minute time intervals and the pickup time and dropoff time are transformed into time interval index. 
Grids shown as bold red squares cover the CBD of NYC, which is the area between 19th street and 59th street in Manhattan. The proposed congestion pricing is applied to vehicles cross the red square into the CBD. 

\begin{figure}[H]
	\centering
	\includegraphics[width=0.99\linewidth,height=0.45\textheight,keepaspectratio]{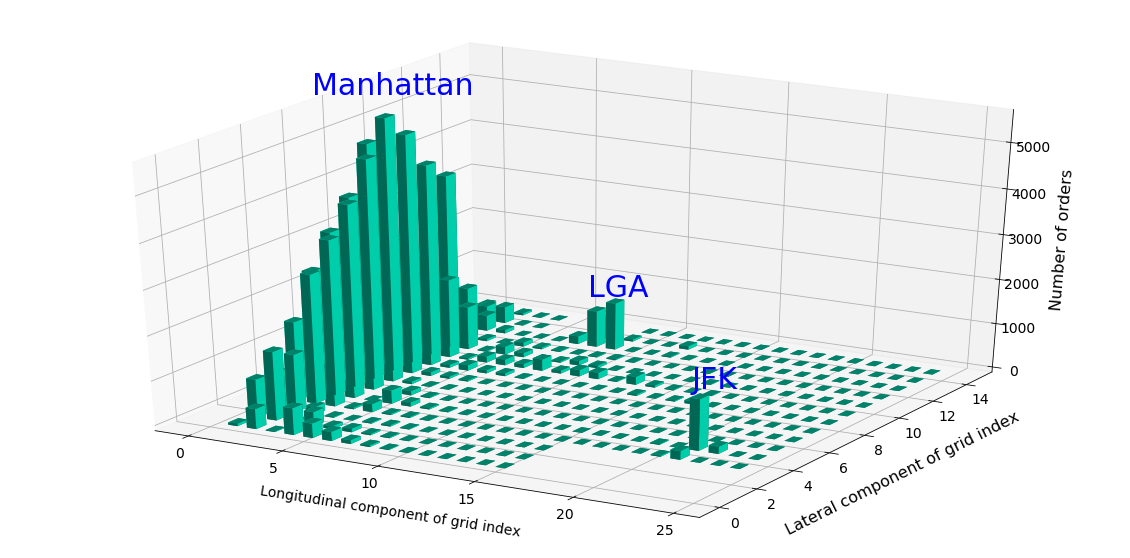}
	\centering 
	\caption{Spatial distribution of taxi orders during evening peak}
	\label{fig:spatial_dist_taxi_orders}
\end{figure}

Figure~(\ref{fig:spatial_dist_taxi_orders}) presents the spatial distribution of taxi orders (pickup) during evening peak. It can be seen that the majority of taxi orders emerge in Manhattan, especially in the CBD. There are two local hotspots near two airports. 

NYC subway turnstile data is also publicly accessible (http://web.mta.info/developers/turnstile.html). A sample of the turnstile data is listed in Table~(\ref{tab:turnstile_sample}). 
These two rows show that the reading of entries for turnstile ID (A002, R051, 02-00-00) is 4,593,637 at 4 PM and 4,594,523 at 8 PM on 05/01/2014. 
Taking the difference between two readings yields the net entries at this turnstile during the 4-hour time interval, i.e., 4,594,523 - 4,593,637 =  886. Similarly, we can calculate net entries and net exits for each turnstile. Net entries and net exits of a grid are then calculated by summing up the net entries and net exits of all turnstiles in that grid, respectively. 

\begin{table}[H]
	\centering\caption{Turnstile data sample}
	\label{tab:turnstile_sample}
	\begin{tabular}{| p{120pt}  | p{50pt} | p{35pt} | p{50pt} | p{50pt} | } 
		\hline
		Turnstile ID & Date & Time & Entries & Exits   \\ \hline
		(A002, R051, 02-00-00) & 05/01/2014 & 16:00:00	& 4,593,637	& 1,564,283 \\ \hline
		(A002, R051, 02-00-00) & 05/01/2014 & 20:00:00	& 4,594,523  & 1,564,348 \\ \hline
	\end{tabular}
\end{table}

\subsubsection{Lower level MARL setup}





	To be concrete, we now provide the setup of the lower level MARL.
	\begin{itemize}
		\item Observation. The observation space for driver $i$ consists of the grid index and current time, i.e., $o_i = (l_i, t)$. One-hot encoding is used for both grid index and time. 
		Yellow and green taxis share the same observation space.
		\item Action. The action space for driver $i$ is to enter one of neighboring grids or to stay at the current grid. Due to a grid world setup with square grids, the action space is thus 5-dimensional (using one-hot encoding). For example, driver $i$ currently at grid $l_i = (5,4)$ chooses action $a_i = \rightarrow$ (i.e., $(1,0)$ in the 2-dimensional grid world setup). The driver will arrive in grid $(6,4)$ to search for passengers. In the boundary grids, we still allow drivers to choose any action from the 5-dimensional action space. If a driver chooses an action that will lead her out of the study area after taking the action, the driver receives a large penalty (i.e., a large negative reward) and is forced to stay in the current grid. Thus drivers will not choose an action that will lead them out of study area after training. 
		\item Reward. The reward of each state transition for driver $i$ is her monetary return, i.e., $$r = \text{fare} - \text{toll charge},$$ where
		$$
		\text{toll charge} =
		\begin{cases}
		\alpha & \text{if driver goes from outside the CBD into the CBD},\\
		0 & \text{otherwise.}
		\end{cases}   
		$$
	\end{itemize}

\textbf{Multiclass MARL.}
The NYC taxi market contains two types of taxicabs, namely yellow taxis and green taxis. They are different because yellow taxis can go and pick up passengers anywhere while green taxis are not allowed to pick up passengers in Manhattan below East 96th Street and West 110th Street and at two airports, namely LaGuardia airport (LGA) and John F. Kennedy airport (JFK).  Therefore we need to model them differently in the lower level MARL. 
To incorporate these two classes of agents into MARL, we create two actors and critics. 
All the yellow taxis share one actor (i.e., a policy network) and one critic (i.e., value network), and green taxis share the other actor and the critic. 
In the actor-critic algorithm demonstrated in Fig~\ref{fig:ac_mean}, both yellow and green agents interact with the same environment. 
They have the same observation space and action space. In other words, for both yellow and green taxis, its observation consists of the grid index and current time, and its action is to enter one of neighboring grids or to stay in the current grid. In addition, both of them aim to maximize their cumulative monetary return. 

The key difference is, green taxis can drop off and search for passengers in those restricted areas (i.e., Manhattan below East 96th Street and West 110th Street and two airports), they can not pick up passengers there. From the modeling perspective, the environment will not assign orders to green taxis in restricted areas. This restriction discourages green taxis from searching for passengers or taking passengers to the restricted areas. Accordingly, the policy for green taxis is expected to be different from that of yellow taxis. 
Yellow taxis thus only compete among themselves in the restricted areas, 
while outside restricted areas, yellow and green taxis not only compete within the same type but also compete with the other taxi type. 


\subsubsection{Upper level objective function of city planners}\label{sec:obj}

As previously mentioned, the objective function of city planners consists of two components, namely, number of vehicles in the CBD and the crowdedness of the public transit. Now we formally define these two components based on NYC taxi data and subway turnstile data. 

The first component is defined as the percentage of taxis in the CBD, i.e, the ratio of the number of taxis in the CBD to the total number of taxis. For each time step, we calculate one value of the percentage. We then take the average of the percentages across all time steps as the overall percentage of taxis in the CBD. Hereafter we call this PTC (\textbf{p}ercentage of \textbf{t}axis in the \textbf{C}BD). PTC decreases with toll charge because fewer vacant taxis enter the CBD with a higher toll charge. 

\begin{equation*}
PTC = \dfrac{ \sum_{t \in \{1,2,\cdots,T\}} \dfrac{\sum_{l \in CBD} \text{num\_taxis}_{tl}}{\text{total\_num\_taxis}} }{T},
\end{equation*}
where $\text{num\_taxis}_{tl}$ is the number of taxis in grid $l$ at time step $t$ and $\text{total\_num\_taxis}$ is the total number of taxis. 


The crowdedness of the subway system in each grid is further decomposed into two parts, namely, the entry crowdedness which is related with the net entries into the subway system within the grid, and the exit crowdedness which is related with the net exits from the subway system within the grid. After imposing a toll charge on taxis entering the CBD, the crowdedness of the subway system increases due to the unserviced taxi orders. Here we assume that travel demand stemming from the unserviced taxi orders goes to the subway system. 
For a grid $l$, we count the number of passengers of unserviced taxi orders with its origin inside the grid and call this quantity the additional entry into the subway system. We then take the ratio of the additional entry to the net entries within the grid as the increase in the entry crowdedness in grid $l$, denoted as $ICS_{l}^{entry}$ where $ICS$ stands for ``an \emph{{i}ncrease in \textbf{c}rowdedness of the \textbf{s}ubway}". Mathematically, 
\begin{equation*}
ICS_{l}^{entry} = \dfrac{ \sum_{\text{order} \in \text{unserviced orders}}  \mathbbm{1}_{\text{origin}_{\text{order}} \in l} \times \text{num\_passengers}_{\text{order}}}{\text{net\_enteries}_l}.
\end{equation*}
where $\text{origin}_{\text{order}}$ and $\text{num\_passengers}_{\text{order}}$ denote the origin and number of passengers of the order, respectively. Similarly, we can calculate the increase in the exit crowdedness in grid $l$ as
\begin{equation*}
	ICS_{l}^{exit} = \dfrac{ \sum_{\text{order} \in \text{unserviced orders}}  \mathbbm{1}_{\text{destination}_{\text{order}} \in l} \times \text{num\_passengers}_{\text{order}}}{\text{net\_enteries}_l}.
\end{equation*}
Taking the average of the increase in the entry crowdedness and the increase in the exit crowdedness yields the increase in the crowdedness of the grid, namely, 
$$
ICS_l = \dfrac{ICS_{l}^{entry} + ICS_{l}^{exit}}{2}.
$$
Among the overall 337 grids, we focus on the top $m$ grids in terms of crowdedness. Thus the overall ICS is calculated as
$$
ICS = \dfrac{\sum_{l \in \text{top m grids}}ICS_{l}}{m},
$$
where $m=20$ in this case study. 
ICS increases with the toll charge because more passengers are directed to the subway system with a higher toll charge.





From the perspective of city planners, both PTC and ICS are expected to be small. These two components, however, are competing against each other. With a small toll charge, ICS is small but PTC is large; while with a large toll charge, PTC becomes smaller but ICS gets larger. Therefore city planners need to maintain some balance between these two components. Here we use a weighted average of these two components as the objective of city planners. To ensure maximization, we add a minus sign:
\begin{equation}
f = -[w \times \text{PTC} + (1-w) \times \text{ICS}]
\label{eqn:large_objective}
\end{equation}
where $w \in [0,1]$ is the weight for $\text{PTC}$. In this case study, we set $w = \frac{1}{5}$ considering the difference in the magnitude of two components. We also conduct sensitivity analysis to investigate the impact of $w$ on the final result. 

\subsubsection{Results}\label{sec:res}

On weekdays, there are on average around 100,000 taxi orders during the evening peak. According to Wikipedia (https://en.wikipedia.org/wiki/Taxicabs\_of\_New\_York\_City), there are around 13,000 yellow ``medallion" taxicabs in NYC. Considering that some drivers do not work during the evening peak and some drivers work outside the grid world, we thus set the number of yellow agents in MARL as 12,000. The number of green agents is set to 5,000 considering that there were in total around 6,000 green taxi drivers in 2015 and some of them may not work during the evening peak. 




	In the mean field actor-critic algorithm, the critic is parameterized by an MLP with four hidden layers $(256,128,64,32)$, and the actor is parameterized by an MLP with four hidden layers $(128,64,32,16)$.  Other hyperparameters remain the same. 

\begin{figure}[H]
	\centering 
	\subfloat[$\alpha=0$]{\includegraphics[scale=.5]{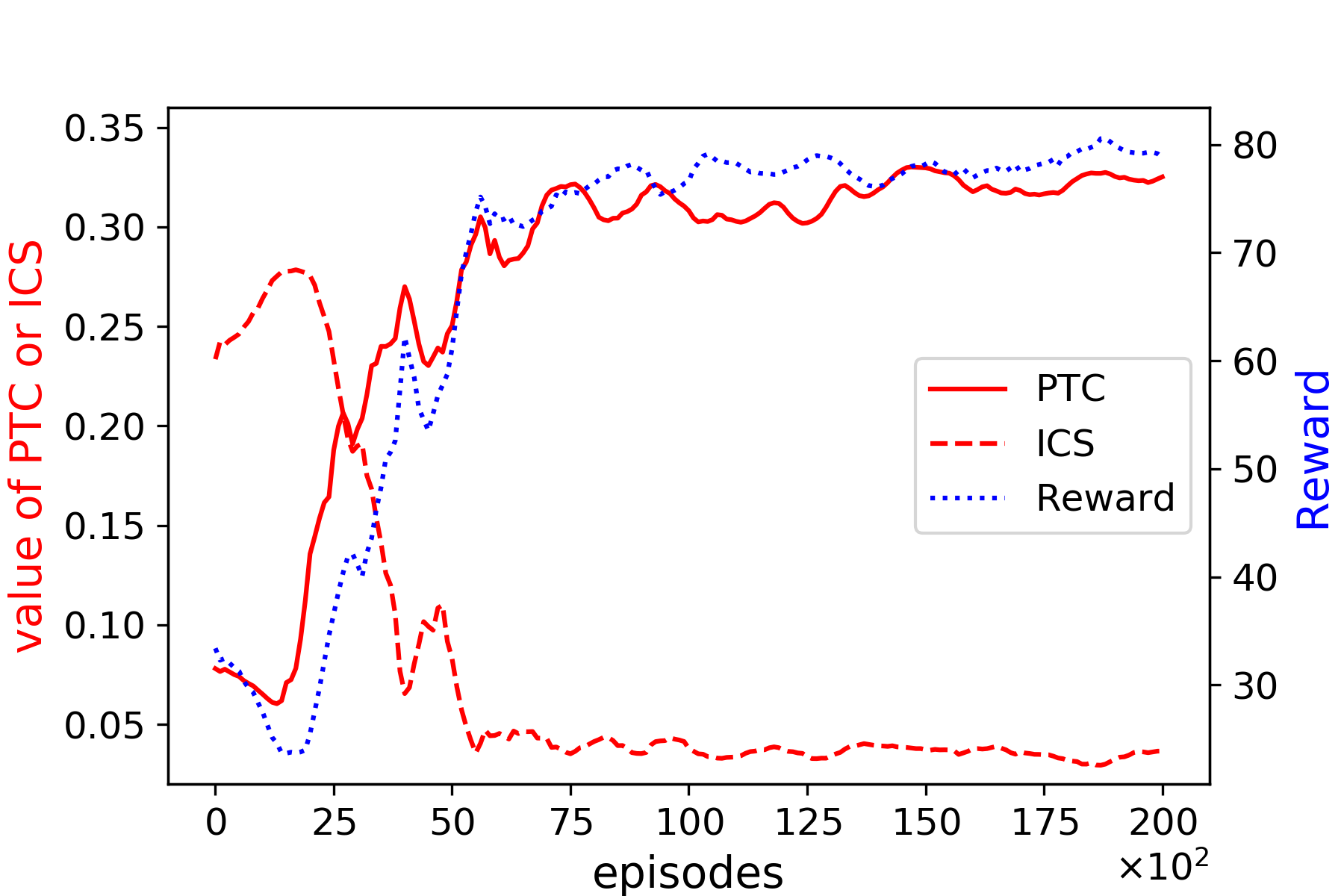}\label{subfig:alpha0_convergence}} ~
	\subfloat[$\alpha=5.1$]{\includegraphics[scale=.5]{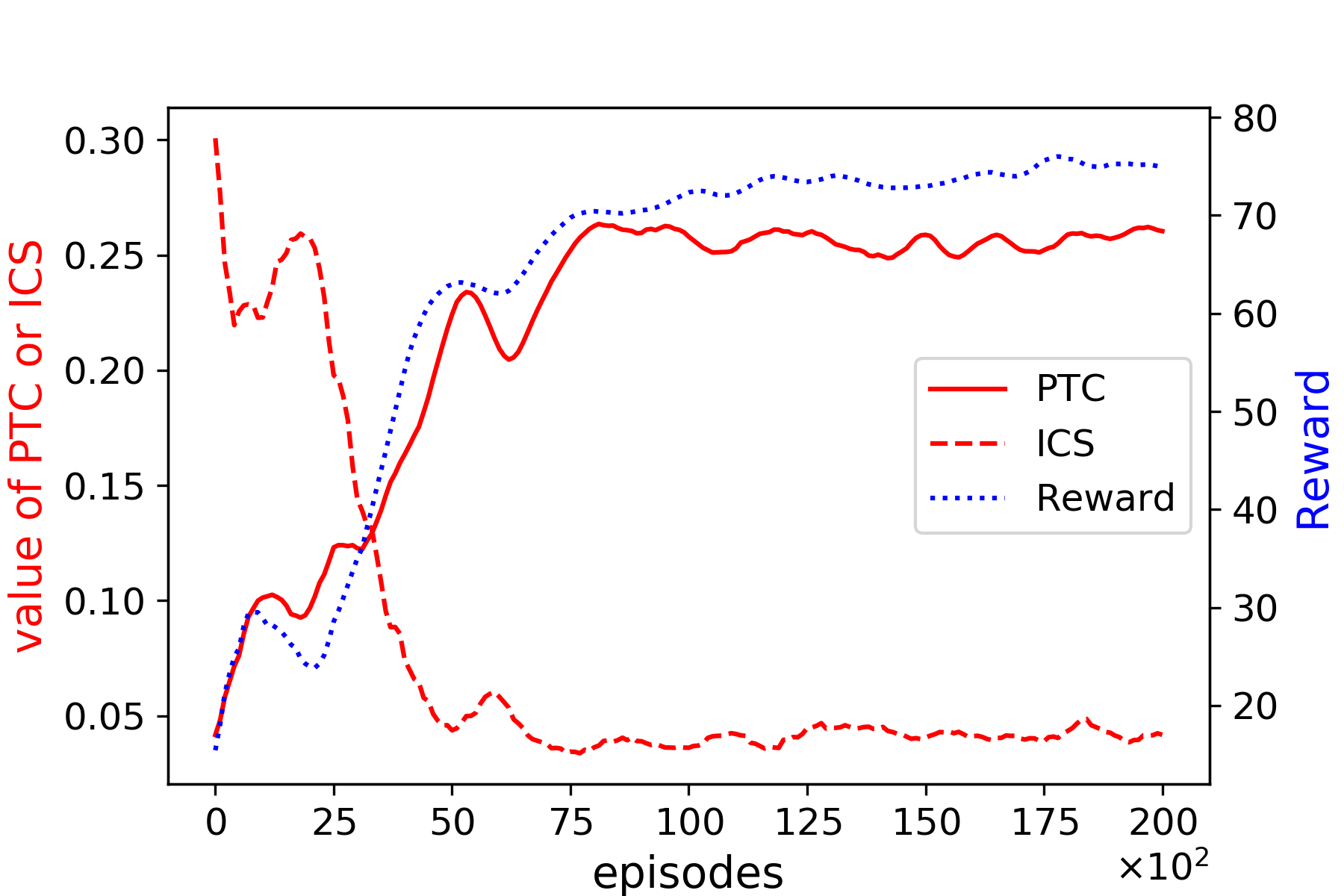}\label{subfig:alpha5_convergence}}
	\caption{Convergence of the lower level MARL}
	\label{fig:convergence_large} 
\end{figure}

For a bilevel optimization problem, first we need to check the convergence of the lower level. As an example to validate the convergence, PTC, ICS, and average cumulative reward (i.e., Reward in Figure~(\ref{fig:convergence_large})) of all agents versus number of episodes are presented in Figure~(\ref{fig:convergence_large}) with $\alpha = 0$ and $\alpha=5.1$. 
In both scenarios, despite some bouncing back and forth due to the initial random exploration of agents, PTC increases fast during the first 7,500 episodes where agents explore the environment and learn towards the optimal policy. Increase in PTC indicates that agents are gradually moving into the CBD to search for passengers, which is as expected.  After 15,000 episodes, PTC reaches its converged value of $0.32$ in the first scenario with $\alpha=0$ and $0.26$ in the second scenario with $\alpha=5.1$, respectively. 
Conversely, ICS decreases during the first 5,000 episodes, meaning that agents are gradually searching in right places so that more passengers requests are fulfilled. ICS reaches its converged value of $0.035$ after 5,000 episodes in the first scenario with $\alpha=0$ and $0.041$ after 7,500 episodes in the second scenario with $\alpha=5.1$, respectively.  
The average cumulative reward reaches its final converged value of $79.3$ in the first scenario with $\alpha=0$ and $75.2$ in the second scenario with $\alpha=5.1$, respectively. 
All three metrics gradually converge after 12,500 episodes when agents mainly exploit the knowledge they have gained through their previous explorations. The computation time for running the mean field actor-critic algorithm until convergence of the lower level MARL is around 38 hours on a virtual machine with 1 NVIDIA Tesla V100 GPU and 8 standard CPUs on Google Cloud Platform. 




\begin{figure}[H]
	\centering 
	\subfloat[Posterior probability distribution of $f$ conditioned on the initial five evaluated points]{\includegraphics[scale=.5]{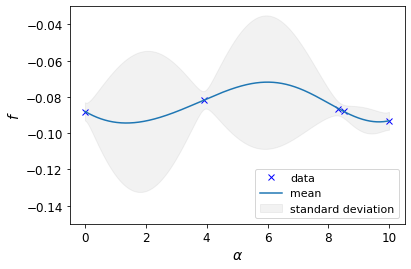}\label{subfig:BO_start_large}} ~
	\subfloat[Acquisition function ($UCB$)]{\includegraphics[scale=.5]{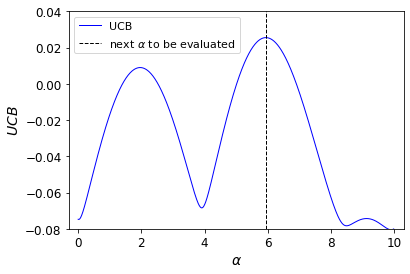}\label{subfig:acqui_large}}
	\caption{Posterior probability distribution and acquisition function at iteration $0$}
	\label{fig:BO_large} 
\end{figure}


With the validated convergence of the lower level MARL, we use BO to solve the bilevel problem. 
In total we evaluate five $\alpha$'s as the starting point of BO. Figure~(\ref{subfig:BO_start_large}) plots the mean and standard deviation of the posterior probability distribution of $f$ conditioned on these five evaluated points. As one can see, the standard deviation is small around the points that have been evaluated and is large at locations where we do not have any data. The acquisition function shown in Figure~(\ref{subfig:acqui_large}) reveals that the next $\alpha$ to be evaluated is $5.95$. According to Figure~(\ref{subfig:BO_start}), the mean is high and the standard deviation is high around $5.95$, indicating a large acquisition. 

	\begin{figure}[H]
	\centering
	\includegraphics[width=0.99\linewidth,height=0.2\textheight,keepaspectratio]{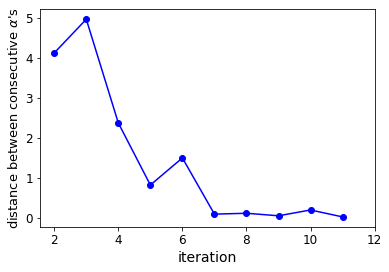}
	\centering 
	\caption{Convergence of BO}
	\label{fig:convergence_bo_large}
\end{figure}

The convergence of BO in this case is defined as choosing 5 consecutive $\alpha$'s with the distance between every two consecutive $\alpha$'s below a threshold of $0.5$. 
The convergence of BO is plotted in Figure~(\ref{fig:convergence_bo_large}).  
We can see that BO initially chooses quite different $\alpha$'s and gradually converges from the $7^{th}$ iteration. 

\begin{figure}[H]
	\centering 
	\subfloat[Posterior probability distribution of $f$ after BO converges]{\includegraphics[scale=.5]{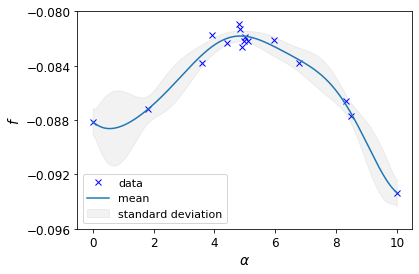}\label{subfig:BO_large_final}} ~
	\subfloat[Two components ]{\includegraphics[scale=.5]{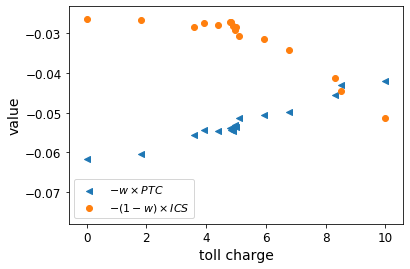}\label{subfig:two_components}}
	\caption{BO result}
	\label{fig:BO_result_large} 
\end{figure}

The resulting posterior probability distribution of $f$ from BO is presented in Figure~(\ref{subfig:BO_large_final}). The optimal toll charge is \$5.1 with an optimal objective around $-0.08$. The objective is $-0.09$ without any toll charge. The objective increases with the toll charge when smaller than $\$5.1$. With a $\$5.1$ toll charge, the objective is $-0.08$, which is $7.9\%$ higher than $-0.09$. The objective decreases if toll charge is increased beyond $\$5.1$. The standard deviation within the range of $[4, 6.2]$ is quite small because we have enough evaluations within the range so the uncertainty is low. As for the range of $[0,2.2]$, $[6.2, 8.3]$, and $[8.3, 10]$, the standard deviation is large, indicating a large uncertainty in these ranges. BO does not choose to evaluate $\alpha$ in these ranges with a large uncertainty because of a low mean value and therefore a small acquisition in these ranges. 
%
The parabolic shape of the objective can be explained by Figure~(\ref{subfig:two_components}). Before the toll charge reaches $\$5.1$, the steady increase in $-w\times PTC$ and the minor decrease in $-(1-w)\times ICS$ push the objective higher with a larger toll charge. After the toll charge is increased beyond $\$5.1$, $-(1-w)\times ICS$ declines faster and suppresses the effect of the increase in $w \times PTC$, resulting in a decrease in the objective. 



\begin{figure}[H]
	\centering
	\includegraphics[width=0.99\linewidth,height=0.2\textheight,keepaspectratio]{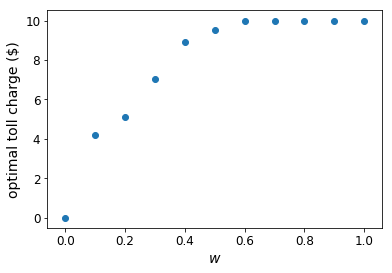}
	\centering 
	\caption{Sensitivity analysis}
	\label{fig:sensitivity}
\end{figure}

	To investigate the impact of the weighting parameter $w$ on the optimal toll charge, sensitivity analysis is conducted by running BO with various values of $w$ and a preset range of $\alpha \in [0, 10]$. The result is presented in Figure~(\ref{fig:sensitivity}).  Recall that city planners aim to minimize both ICS and PTC, and a larger toll charge usually discourages taxis from entering the CBD, resulting in a lower PTC and a higher ICS. 
	With $w=0$, city planners put all weights on ICS, and the optimal toll charge is simply zero.  As $w$ increases, city planners gradually put more weights on PTC, resulting in an increase in the optimal toll charge. With $w=\frac{1}{10}$ and $w=\frac{1}{5}$, optimal toll charges are $\$4.2$ and $\$5.1$, respectively. As plotted in Figure~(\ref{subfig:two_components}), the magnitude of $-w\times PTC$ is already larger than that of $-(1-w)\times ICS$ with $w=\frac{1}{5}$. Increasing $w$ further increases the discrepancy between the magnitude of $-w\times PTC$  and that of $-(1-w)\times ICS$. In other words, the magnitude of $-w\times PTC$ is much larger than that of $-(1-w)\times ICS$ when $w > \frac{1}{5}$, especially when $w > \frac{1}{2}$. With $w > \frac{1}{2}$, the optimal toll charge is around $\$10$, which is the upper bound of the preset range, because now city planners focus much more on PTC than ICS. 
	In practice, the selection of the weight depends on how city planners' weighs in subway crowdedness and traffic congestion arising from taxis in CBD. 
	Figure~(\ref{fig:sensitivity}) also provides a guidance to planners if they also need to maintain a reasonable value of the optimal toll.

\begin{figure}[H]
	\centering 
	\subfloat[Percentage of taxis in each grid (grids with percentage lower than 0.001 are omitted)]{\includegraphics[scale=.3]{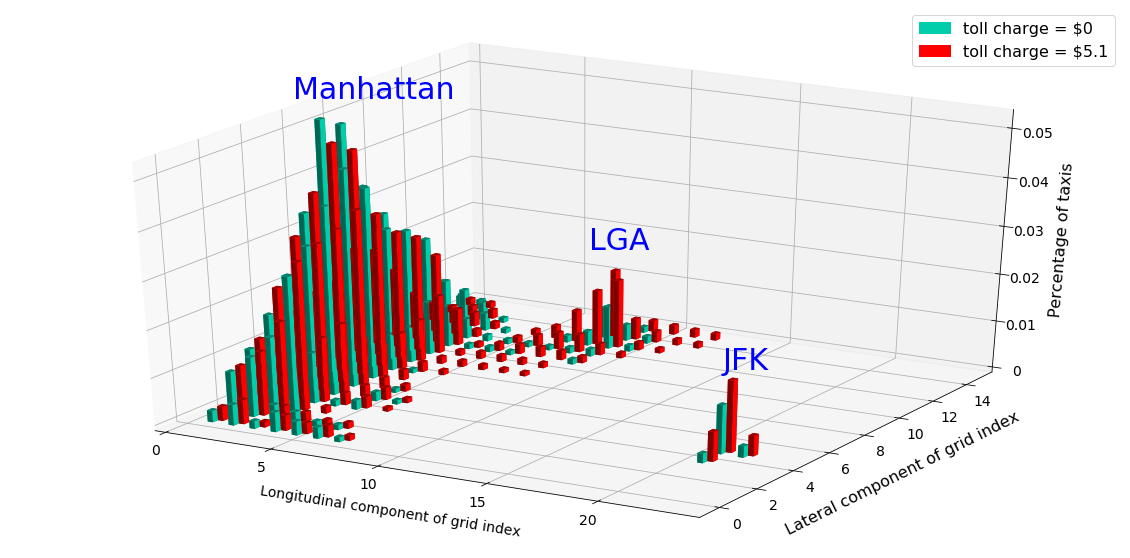}\label{subfig:PTC}} \\
	\subfloat[Increase in crowdedness across the busiest 20 grids ]{\includegraphics[scale=.3]{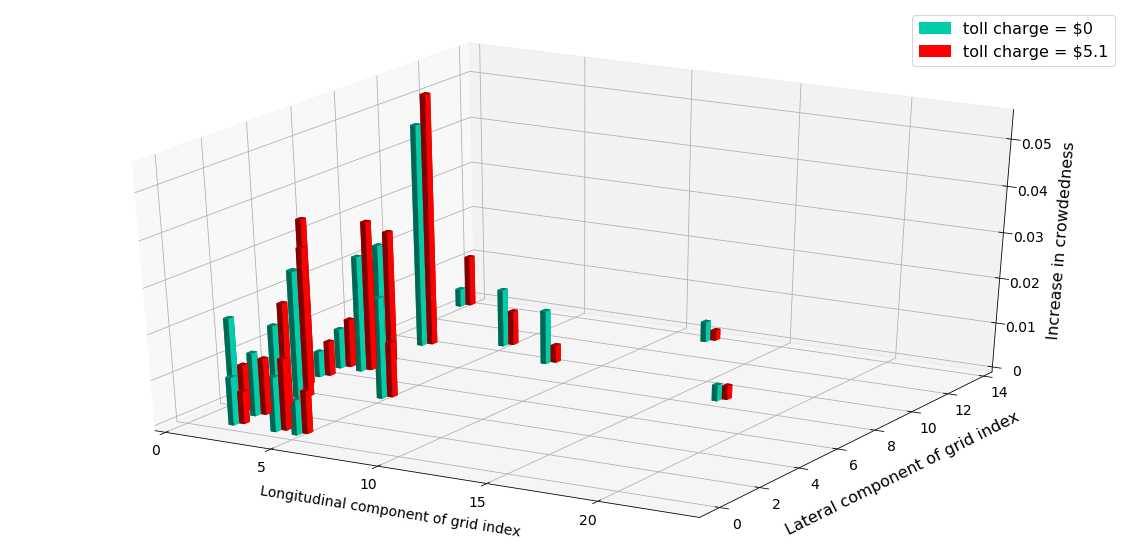}\label{subfig:ICS}}
	\caption{Spatial distribution of percentage of taxis in each grid and increase in crowdedness across the busiest 20 grids}
	\label{fig:two_components_spatial} 
\end{figure}

Figure~(\ref{subfig:PTC}) presents the average percentage of taxis in each grid for two scenarios, namely without any toll charge and with the optimal toll charge. With the optimal toll charge, the percentage of taxis in Manhattan, especially in the CBD, is decreased, while that for two airports are increased. This is as expected because taxi drivers are penalized for entering the CBD vacantly, meaning that CBD now becomes less attractive to taxi drivers. According to the demand distribution shown in Figure~(\ref{fig:spatial_dist_taxi_orders}), two airports become comparatively attractive.  Figure~(\ref{subfig:ICS}) presents the increase in crowdedness across the busiest 20 grids. With the optimal toll charge, the increase in crowdedness in the subway is higher in the CBD, compared to that without any toll charge, because now there are fewer taxis in the CBD and therefore more passengers are directed to the subway system. For grids outside CBD, the increase in crowdedness can be either higher or lower because the crowdedness consists of two components, namely, the entry crowdedness and the exit crowdedness. The increase in entry crowdedness is expected to be lower for grids outside CBD because there are more vacant taxis outside the CBD who is willing to carry passengers into CBD. The increase in exit crowdedness is higher in many grids because more people take subway to arrive in grids outside the CBD.  

\section{Conclusion}\label{sec:conclu}

Noticing the underutilization of taxi resources due to idle taxi drivers' cruising behavior, this study aims to model the multi-driver repositioning task through a mean field multi-agent reinforcement learning approach. A mean field actor-critic algorithm is developed to solve the MARL with a given reward function. The direct application of the mean field actor-critic algorithm is, however, very likely to yield a suboptimal equilibrium from the standpoint of the system. Thus, this study proposes a bilevel optimization with the upper level as a reward design and the lower level as an MAS. The upper level interacts with the lower level by adjusting rewards. 
The bilevel optimization model is applied to two scenarios, namely, e-hailing driver repositioning under service charge and taxi driver repositioning under congestion pricing. In the case of e-hailing driver repositioning, the agreement between the derived optimal control from BO and that from an analytical solution validates the effectiveness of the model.  It is also worth mentioning that the objective of the e-hailing platform is increased by $8.4\%$ using a simple piecewise linear platform service charge. In the case of multiclass taxi driver repositioning, a $\$5.1$ toll charge increases the objective of city planners by $7.9\%$, compared to that without any toll charge. With the optimal toll charge, the number of taxis in the CBD is decreased, indicating a better traffic condition. The crowdedness is increased in the subway stations within the CBD due to fewer taxis. For subway stations outside the CBD, the crowdedness can be either higher or lower depending on the tradeoff between the entry crowdedness and the exit crowdedness. 

The aforementioned two driver-repositioning applications validate the effectiveness of the proposed bilevel optimization model. We stress that the model is general and can be applied to various systems as long as there are two levels in the system and the upper level can affect the lower level through some control. With some optimal control, the performance of the system can be improved, which is beneficial for the urban economy. 

There are some future work that can be done to overcome some limitations of this study.  

\begin{enumerate}
	
	\item In this study, drivers are assumed to be intelligent and perfectly rational. Bounded rationality could be studied to more realistically capture the boundedly rational behavior of real drivers. 
	For example, in our first case study, to maximize ORR, we assume the platform chooses the largest possible value of $\alpha$. There might exist a threshold or an indifference band. When $\alpha$ if larger than such a threshold, drivers would take other actions and lead to a higher system ORR. 
	
	\item 
	Although the difference between yellow taxis and green taxis is considered in the second case study, the heterogeneity among yellow taxis (or green taxis) is neglected due to the homogeneity assumption. A more personalized model capturing the behavioral difference within the same type of agents is left in future research.
	
	\item In the second case study, the demand is modeled separately from the supply without considering the coupling effect between the two. In addition, only two modes, namely taxi and subway are used due to data accessibility. In reality, passengers' mode choice and willingness to pay need to be modeled to capture their impact on the demand, which subsequently affect the distribution of supply.
	
	
\end{enumerate}

\section*{Acknowledgements}
This work is partially sponsored by The Region 2 University Transportation Research Center (UTRC) and the National Science Foundation under CAREER award number CMMI-1943998.

\bibliographystyle{elsarticle-harv}
\bibliography{marl}
\end{document}